\documentclass[10pt,journal]{IEEEtran}

\usepackage[cmex10]{amsmath}

\usepackage[]{cite}
\usepackage{url}

\usepackage{graphicx}
\usepackage{amsmath}
\usepackage{marvosym}
\usepackage{epsfig}
\usepackage{amssymb}
\usepackage{caption}
\usepackage[svgnames,table]{xcolor}
\usepackage{comment}
\usepackage{algorithm}
\usepackage[algo2e,vlined,algoruled,resetcount]{algorithm2e}
\usepackage{enumitem}
\usepackage{multirow}
\usepackage{subfig}
\usepackage[pagebackref=true,breaklinks=true,letterpaper=true,colorlinks,bookmarks=false]{hyperref}

\usepackage{cases}
\usepackage{wrapfig}

\usepackage{textcomp}

\makeatletter
\DeclareRobustCommand\onedot{\futurelet\@let@token\@onedot}
\def\@onedot{\ifx\@let@token.\else.\null\fi\xspace}

\def\eg{\emph{e.g}\onedot} 
\def\ie{\emph{i.e}\onedot} 
 
\def\etc{\emph{etc}\onedot} 
 
\def\etal{\emph{et al}\onedot}

\makeatother

\def\KBName{FVQA\xspace}
\def\ModParms{{\boldsymbol \theta}}

\def\bluetext{}
\def\bluett{}
\def\bluettt{\textcolor{black}}

\begin{document}
\sloppy

\title{\KBName: Fact-based Visual Question Answering %
}
\author{Peng Wang$^*$,
        Qi Wu$^*$,
        Chunhua Shen,
        Anthony Dick,
        Anton~van~den~Hengel
        \IEEEcompsocitemizethanks{\IEEEcompsocthanksitem
P. Wang is with Northwestern Polytechnical University, China, and The University of Adelaide, Australia.
Q. Wu, C. Shen, A. Dick and A. van den Hengel are with The University of Adelaide, Australia.
(E-mail: peng.wang@nwpu.edu.cn, \{qi.wu01, chunhua.shen, anthony.dick, anton.vandenhengel\}@adelaide.edu.au).
  Correspondence should be addressed to C. Shen.
$^*$The first two authors contributed to this work equally.
}
}

\maketitle

\begin{abstract}

Visual Question Answering (VQA) has attracted much attention in both computer vision and natural language processing communities, not least because it offers insight into the relationships between two important sources of information.
Current datasets, and the models built upon them, have focused on questions which are answerable by direct analysis of the question and image alone.  The set of such questions that require no external information to answer is interesting, but very limited.  It excludes questions which require common sense, or basic factual knowledge to answer, for example.
Here we introduce \KBName (Fact-based VQA), a VQA dataset which requires, and supports, much deeper reasoning.
\KBName primarily contains questions that require external information to answer.
We thus extend a conventional visual question answering dataset, which contains {image-question-answer} triplets, through additional {image-question-answer-supporting fact} tuples.
Each supporting-fact is represented as a structural triplet, such as \texttt{<Cat,CapableOf,ClimbingTrees>}.

We evaluate several baseline models on the \KBName dataset, and describe a novel model which is capable of reasoning about an image on the basis of supporting-facts.

\end{abstract}


\tableofcontents
\clearpage

\section{Introduction}
\label{intro}
Visual Question Answering (VQA) can be seen as a proxy task for evaluating a vision system's capacity for deeper image understanding. It requires elements of image analysis, natural language processing, and a means by which to relate images and text. Distinct from many perceptual visual tasks such as image classification, object detection and recognition~\cite{krizhevsky2012imagenet, lin2014microsoft, deng2009imagenet, simonyan2014very}, however, VQA requires that a method be prepared to answer a question that has never been seen before.  In object detection the set of objects of interest are specified at training time, for example, whereas in VQA the set of questions which may be asked inevitably extend beyond those in the training set.

The set of questions that a VQA method is able to answer are one of its key features, and limitations. Asking a method a question that is outside its scope will lead to a failure to answer, or worse, to a random answer.
Much of the existing VQA effort has been focused on questions which can be answered by the direct analysis of the question and image, on the basis of a large training set~\cite{antol2015vqa,malinowski2014towards,gao2015you,Yu_2015_ICCV,ren2015image,zhu2015visual7w}.
This is a restricted set of questions, which require only relatively shallow image understanding to answer.  It is possible, for example, to answer `How many giraffes are in the image?' without understanding any non-visual knowledge about giraffes.

The number of VQA datasets available has grown as the field progresses~\cite{antol2015vqa,malinowski2014towards,gao2015you,Yu_2015_ICCV,ren2015image,zhu2015visual7w}. They have contributed valuable large-scale data for training neural-network based VQA models and introduced various question types, and tasks, from global association between QA pairs and images~\cite{antol2015vqa,malinowski2014towards,ren2015image} to grounded QA in image regions~\cite{zhu2015visual7w}; from free-from answer generation~\cite{antol2015vqa,gao2015you,ren2015image,zhu2015visual7w} to multiple-choice picking~\cite{antol2015vqa,malinowski2014towards} and blank filling~\cite{Yu_2015_ICCV}.
For example, the questions defined in DAQUAR~\cite{malinowski2014towards} are almost exclusively
``Visual'' questions, referring to ``color'', ``number'' and ``physical location of the object''. In the COCO-QA dataset~\cite{ren2015image}, questions are generated automatically from image captions which describe the major visible content of the image.

\begin{figure}[t!]
\begin{center}
	\includegraphics[width=0.95\columnwidth]{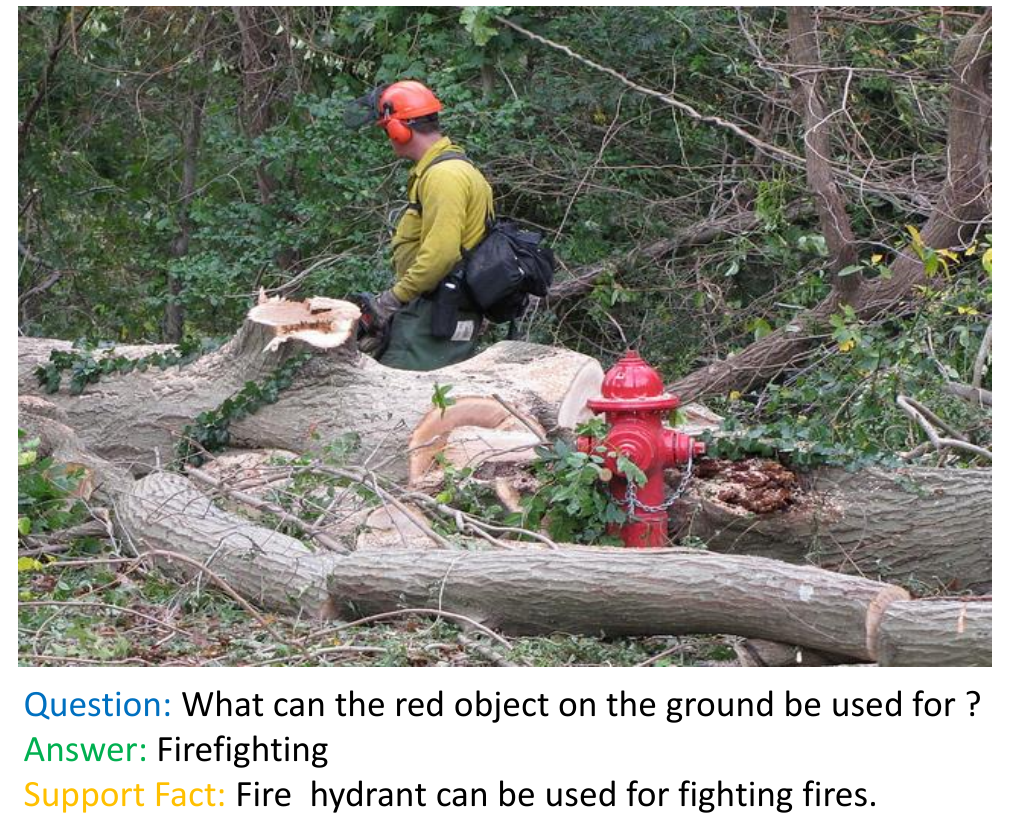}
\end{center}
\vspace{-4pt}
\caption{An example visual-based question from our \KBName dataset that requires both visual and common-sense knowledge to answer. The answer and mined knowledge are generated by our proposed method.}
\label{fig:t_figure}
\vspace{-5pt}
\end{figure}

\bluetext{
The VQA dataset in~\cite{antol2015vqa}, for example, has been very well studied, yet only 5.5\% of questions require adult-level (18+) knowledge (28.4\% and 11.2\% questions require older child (9-12) and teenager (13-17) knowledge). }
This limitation means that this is not a truly ``AI-complete'' problem, because this is not a realistic test for human beings. Humans inevitably use their knowledge to answer questions, even visual ones.
For example, to answer the question given in  Fig. \ref{fig:t_figure}, one not only needs to visually recognize the `red object' as a `fire hydrant', but also to know that `{\em a fire hydrant can be used for fighting fires}'.

Developing methods that are capable of deeper image understanding demands a more challenging set of questions.
We consider here the set of questions which may be answered on the basis of an external source of information, such as Wikipedia.
This reflects our belief that reference to an external source of knowledge is essential to general VQA.  This belief is based on the observation that the number of $\{$\textit{image-question-answer}$\}$ training examples that would be required to provide the background information necessary to answer general questions about images would be completely prohibitive.
The number of concepts that would need to be illustrated is too high, and scales combinatorially.

In contrast to previous VQA datasets which only contain \textit{question-answer} pairs for an image, we additionally provide a \textit{supporting-fact} for each \textit{question-answer} pair, as shown in Fig.~\ref{fig:t_figure}. The \textit{\bf supporting-fact} is a structural representation of knowledge that is stored in external KBs and indispensable for answering a given visual question.
For example, given an image with a cat and a dog, and the question  \texttt{`Which animal in the image is able to climb trees?'},  the answer is \texttt{`cat'}. The required supporting-fact for answering this question is \texttt{<Cat,CapableOf,ClimbingTrees>}, which is extracted from an existing knowledge base. By providing supporting-facts, the dataset supports answering complex questions, even if all of the information required to answer the question is not depicted in the image. Moreover, it supports explicit reasoning in visual question answering, \ie, it gives an indication  as to how a method might derive an answer. This information can be used in answer inference, to search for other appropriate facts, or to evaluate answers which include an inference chain.

In demonstrating the value of the dataset in driving deeper levels of image understanding in VQA, we
examine on our \KBName dataset 
the performance of the state-of-the-art RNN (Recurrent Neural Network) based approaches~\cite{antol2015vqa,malinowski2014towards,ren2015image}.  We find that there are a number of limitations with these approaches. Firstly, there is no explicit reasoning process in these methods. This means that it is impossible to tell whether the method is answering the question based on image information or merely the prevalence of a particular answer in the training set. The second problem is that, because the model is trained on individual question-answer pairs, the range of questions that can be accurately answered is limited. It can only answer questions about concepts that have been observed in the training set, but there are millions of possible concepts and hundreds of millions relationships between them. %

Our main contributions are as follows.
A new VQA dataset (\KBName) with additional supporting-facts is introduced in Sec.~\ref{dataset}, which requires and supports deeper reasoning.
In response to the observed limitations of RNN-based approaches, we propose a method which is based on explicit reasoning about the visual concepts detected from images, in Sec.~\ref{method}. The proposed method first detects relevant content in the image, and relates it to information available in a pre-constructed knowledge base (we combine several publicly available large-scale knowledge bases). A natural language question is then automatically classified and mapped to a query which runs over the combined image and knowledge base information. The response of the query leads to the supporting-fact, which is then processed so as to form the final answer to the question. 
\bluettt{Our approach achieves the Top-1 accuracy of $56.91\%$, outperforming existing baseline VQA methods.}

\section{Related Work}
\label{relwork}

\subsection{Visual Question Answering Datasets}

Several datasets designed for Visual Question Answering have been proposed. The DAQUAR \cite{malinowski2014towards}
dataset is the first small benchmark dataset built upon indoor scene RGB-D images, which is mostly composed of questions requiring only visual knowledge. Most of the other datasets \cite{antol2015vqa,gao2015you,Yu_2015_ICCV,ren2015image,zhu2015visual7w} represent question-answer pairs for Microsoft COCO images \cite{lin2014microsoft}, either generated automatically by NLP tools \cite{ren2015image} or written by human workers \cite{antol2015vqa,gao2015you}. \bluetext{The Visual Genome dataset~\cite{krishnavisualgenome} contains 1.7 million questions, which are asked by human workers based on region descriptions. }The MadLibs dataset \cite{Yu_2015_ICCV} provides a large number of template based text descriptions of images, which are used to answer multiple choice questions about the images. Visual 7W  \cite{zhu2015visual7w} established a semantic link between textual descriptions and image regions by object-level grounding and the questions are asked based on groundings.

\subsection{Visual Question Answering Methods}
Malinowski \etal \cite{malinowski2014multi}
were
the first to study the VQA problem. They proposed a method that combines image segmentation and semantic parsing with a Bayesian approach to sample from nearest neighbors in the training set. This approach requires human defined relationships, which are inevitably dataset-specific. Tu \etal~\cite{tu2014joint} built a query answering system based on a joint parse graph from text and videos. Geman \etal~\cite{geman2015visual} proposed an automatic `query generator' that is trained on annotated images and produces a sequence of binary questions from any given test image.

The current dominant trend within VQA is to combine convolutional neural networks and recurrent neural networks to learn the mapping from input images and questions, to answers. Both Gao \etal~\cite{gao2015you} and Malinowski \etal \cite{malinowski2015ask} used RNNs to encode the question and generate the answer.  Whereas Gao \etal~\cite{gao2015you} used two networks, a separate encoder and decoder, Malinowski \etal \cite{malinowski2015ask} used a single network for both encoding and decoding. Ren \etal \cite{ren2015image} focused on questions with a single-word answer and formulated the task as a classification problem using an LSTM (Long Short Term Memory) network. Inspired by Xu \etal \cite{xu2015show} who encoded visual attention in image captioning, authors of \cite{zhu2015visual7w,Chen2015ABC,Jiang2015compositional,andreas2015deep,yang2015stacked} proposed to use spatial attention to help answer visual questions. 
\bluettt{Most of existing methods formulated the VQA as a classification problem and restrict that the answer only can be drawn from a fixed answer space.} 
In other words, they cannot generate open-ended answers. 
Zhu~\etal~\cite{zhu2015uncovering} investigated the video question answering problem using `fill-in-the-blank' questions. However, either an LSTM or a GRU (Gated Recurrent Unit) is still applied in these methods to model the questions. Irrespective of the finer details, we refer to them as the \bluettt{RNN} approaches.

\subsection{Knowledge-bases and VQA}

Answering general questions posed by humans about images inevitably requires reference to information not contained in the image itself. To an extent this information may be provided by an existing training set such as ImageNet~\cite{deng2009imagenet}, or Microsoft COCO~\cite{lin2014microsoft} as class labels or image captions.
There are a number of forms of such auxiliary information, including, for instance, question/answer pairs
which refer to objects that are not depicted (\eg, which reference people waiting for a train, when the train is not visible in the image) and provide external knowledge that cannot be derived directly from the image (\eg, the person depicted is Mona Lisa).

Large-scale {\em structured} Knowledge Bases
(KBs)~\cite{auer2007dbpedia,banko2007open,bollacker2008freebase,carlson2010toward,chen2013neil,mahdisoltani2014yago3,vrandevcic2014wikidata}
in contrast, offer an explicit, and typically larger-scale, representation of
such external information. In structured KBs, knowledge is typically represented by a large number of triples of the form \texttt{(arg1,rel,arg2)}, which we refer to as {\bf Facts} in this paper.
\texttt{arg1} and \texttt{arg2} denote two {\bf Concepts} in the KB,
each describing a concrete or abstract entity with specific characteristics.
\texttt{rel} represents a specific {\bf Relationship} between them.
A collection of such triples form a large interlinked graph. Such triples are often described according to a Resource Description Framework~\cite{rdf2014resource} (RDF) specification, and housed in a relational database management system (RDBMS), or triple-store, which allows queries over the data.
The information in KBs can be accessed efficiently using a query language.
In this work we use SPARQL Protocol~\cite{prud2008sparql} to query the OpenLink Virtuoso~\cite{erling2012virtuoso} RDBMS.

Large-scale structured KBs are constructed either by manual annotation
(\eg, DBpedia~\cite{auer2007dbpedia}, Freebase~\cite{bollacker2008freebase} and Wikidata~\cite{vrandevcic2014wikidata}), or by automatic extraction from unstructured/semi-structured data (\eg, YAGO~\cite{hoffart2013yago2,mahdisoltani2014yago3}, OpenIE~\cite{banko2007open,etzioni2011open,fader2011identifying}, NELL~\cite{carlson2010toward}, NEIL~\cite{chen2013neil},
WebChild~\cite{tandon2014acquiring}, ConceptNet~\cite{liu2004conceptnet}). The KB that we use here is the combination of DBpedia, WebChild and ConceptNet, which contains structured information extracted from Wikipedia and unstructured online articles.

\bluett{
In the NLP and AI communities, there is an increasing interest in the problem of natural language question answering over structured KBs (referred to as KB-QA). %
KB-QA approaches can be divided into two categories: {\em semantic parsing} methods~\cite{zettlemoyer2012learning,zettlemoyer2009learning,berant2013semantic,cai2013large,liang2013learning,kwiatkowski2013scaling,berant2014semantic,fader2014open,yih2015semantic,reddy2016transforming,xiao2016sequence} that
try to translate questions into accurate logical expressions and then map to KB queries, and
{\em information retrieval} methods~\cite{unger2012template,kolomiyets2011survey,yao2014information,bordes2014question,bordes2014open,dong2015question,bordes2015large} that coarsely retrieve a set of answer candidates and then perform ranking.
Similar to information retrieval methods, the QA systems using memory networks~\cite{weston2014memory,sukhbaatar2015end,bordes2015large,kumar2016ask,xiong2016dynamic} record
supporting-fact candidates in a memory module that can be read and written to.
The memory networks are trained to find the supporting-fact that leads to the answer by performing lookups in memory.
The ranking of stored facts can be implemented by measuring the similarity between facts and the question 
using attention mechanisms.
}

\bluett{
VQA systems that exploit KBs are still relatively rare.
Our proposed approach, which generates KB queries to obtain answers, is similar to semantic-parsing based methods.  
Alternatively, information-retrieval based methods, or memory networks, can be also used for the problem
posed in this paper. We leave this for future work.
The work in \cite{zettlemoyer2012learning} maps natural language queries to structured expressions in the lambda calculus.
Similarly, our VQA dataset also provides the supporting fact for a visual question, which, however, is not necessarily an equivalent translation of the given question.
}

 Zhu \etal~\cite{zhu2015building} used a KB and RDBMS to answer image-based queries. However, in contrast to our approach, they build a KB for the purpose, using an MRF model, with image features and scene/attribute/affordance labels as nodes. The  links between nodes represent mutual compatibility relationships. The KB thus relates specific images to specified image-based quantities, which are all that exists in the database schema. This prohibits question answering that relies on general knowledge about the world. Most recently, Wu \etal \cite{wu2015ask} encoded the text mined from DBpedia to a vector with the Word2Vec model which they combined with visual features to generate answers using an LSTM model. However, their proposed method only extracts discrete pieces of text from the knowledge base, thus ignoring the power of its structural representation. Neither \cite{zhu2015building} nor \cite{wu2015ask} are capable of explicit reasoning, in contrast to the method we propose here.

The approach closest to that we propose here is that of Wang \etal~\cite{wang2015explicit}, as it is capable of reasoning about an image based on information extracted from a knowledge base. However, their method largely relies on the pre-defined template, which only accepts questions in a pre-defined format. Our method here does not suffer this constraint. Moreover, their proposed model used only a single manually annotated knowledge source whereas the method here uses this plus two additional automatically-learned knowledge bases.
This is critical because manually constructing such KBs does not scale well, and using automatically generated KBs thus enables the proposed method to answer more general questions.

\bluett{
Krishnamurthy and Kollar~\cite{krishnamurthy2013jointly} 
proposed a semantic-parsing based method to find the grounding of a natural language query in a physical environment (such as image scene or geographical data).
Similar to our approach, a structured, closed-domain KB is firstly constructed for a specific environment, 
then the natural language query is parsed into a logical form and its grounding is predicted.
However, our approach differs from~\cite{krishnamurthy2013jointly} in that it focuses on 
visual questions requiring support from both physical environment and external commonsense knowledge,
which predicts over a much larger, multi-domain KB. }

\bluett{
	Similarly to our approach, 
	Narasimhan \etal \cite{narasimhan2016improving} also query data from external sources when
	information in the existing data is incomplete.
	The work in \cite{narasimhan2016improving} performs queries over unstructured Web articles, while ours 
	performs queries over structured relational KBs.
	Furthermore, the query templates used in \cite{narasimhan2016improving} are much simpler: only the article’s title
	needs to be replaced.}

\section{Creating the \KBName Dataset}
\label{dataset}
Different from previous VQA datasets \cite{antol2015vqa,gao2015you,Yu_2015_ICCV,ren2015image,zhu2015visual7w} that only ask annotators to provide {\em question-answer} pairs without any restrictions, 
{the questions in our dataset are expected to be answered with support of some commonsense knowledge.}
This means that we cannot simply distribute only images to questioners like others \cite{antol2015vqa,zhu2015visual7w}. 
\bluett{
We need to provide a large number of supporting-facts (commonsense knowledge) which are linked to 
concepts that can be grounded in images (we refer to as {\bf Visual Concepts}). }
We build our own on-line question collection system and allow users to choose images, visual concepts and candidate supporting-facts freely. Then users can ask questions based on their previous choices (all choices will be recorded). We provide users with a tutorial and restrict them to ask questions that only to be answered with both visual concept in the image and the provided external commonsense knowledge. In the following sections, we provide more details about images, visual concepts, knowledge bases and our question collection system and procedures. We also compare with other VQA datasets with data statistics.

\subsection{Images and Visual Concepts}
\label{sec:imgvc}

We sample $2190$ images from the Microsoft COCO~\cite{lin2014microsoft} validation set and ImageNet~\cite{deng2009imagenet} test set for collecting questions. Images from Microsoft COCO can provide more context because they have more complicated scenes. Scenes of ImageNet images are simpler but there are more object categories ($200$ in ImageNet vs. $80$ in Microsoft COCO).

Three types of visual concepts are extracted in this work:

\noindent{\bf $-$ Object:}
Instances of real-world object classes with certain semantic meaning (such as humans, cars, dogs)
are detected by two Fast-RCNN~\cite{girshick2015fast} models that are trained respectively on 
Microsoft COCO $80$-object (train split) and ImageNet $200$-object datasets.
The image attribute model~\cite{qi2015caption} also predicts the existence 
of $92$ objects without localisation information.
Overall, there are $326$ distinct object classes to be extracted.

\noindent{\bf $-$ Scene:}
The image scene (such as office, bedroom, beach, forest) 
information is extracted by combining 
the VGG-16 model trained on MIT Places $205$-class dataset~\cite{zhou2014learning} and
the attribute classifier~\cite{qi2015caption} including $25$ scene classes. 
$221$ distinct scene classes are obtained after the combination.

\noindent{\bf $-$ Action:}
The attribute model~\cite{qi2015caption} provides $24$ classes of actions of humans or animals,
such as walking, jumping, surfing, swimming. 

A full list of extracted visual concepts can be found in the Appendix.
In the next section, these visual concepts are further linked to a variety of external knowledge bases.

\subsection{Knowledge Bases}

\begin{table*}[tpb!]
  \centering
\centering
\footnotesize
{
\resizebox{0.95\linewidth}{!}{
\begin{tabular}{l|p{2.4cm}|c|l}
\hline
& & & \\ [-2ex]
KB &{Relationship} &$\#$Facts &{Examples} \\
& & & \\ [-2ex]
\hline
& & & \\ [-2ex]
DBpedia
&\texttt{Category} 	&$35152$
&\texttt{(\underline{Wii},Category,VideoGameConsole)}  \\
& & & \\ [-2ex]
\hline
& & & \\ [-2ex]
\multirow{10}{*}{ConceptNet}
&\texttt{RelatedTo} 	&$79789$
&\texttt{(\underline{Horse},RelatedTo,\underline{Zebra})}, \texttt{(\underline{Wine},RelatedTo,Goblet)},
\bluett{\texttt{(\underline{Surfing},RelatedTo,Ocean)}} \\
&\texttt{AtLocation} 	&$13683$
&\texttt{(Bikini,AtLocation,\underline{Beach})}, \texttt{(Tap,AtLocation,\underline{Bathroom})} \\
&\texttt{IsA} 		&$6011$
&\texttt{(\underline{Broccoli},IsA,GreenVegetable)} \\
&\texttt{CapableOf} 	&$5837$
&\texttt{(\underline{Monitor},CapableOf,DisplayImages)} \\
&\texttt{UsedFor}   	&$5363$
&\texttt{(\underline{Lighthouse},UsedFor,SignalingDanger)}\\ %
&\texttt{Desires} 	&$3358$
&\texttt{(\underline{Dog},Desires,PlayFrisbee)}, \texttt{(\underline{Bee},Desires,Flower)} \\
&\texttt{HasProperty} 	&$2813$
&\texttt{(\underline{Wedding},HasProperty,Romantic)}  \\
&\texttt{HasA} 	&$1665$
&\texttt{(\underline{Giraffe},HasA,LongTongue)}, \texttt{(\underline{Cat},HasA,Claw)} \\
&\texttt{PartOf} 	&$762$
&\texttt{(RAM,PartOf,\underline{Computer})}, \texttt{(Tail,PartOf,\underline{Zebra})} \\
&\texttt{ReceivesAction} 	&$344$
&\texttt{(\underline{Books},ReceivesAction,bought at a bookshop)} \\
&\texttt{CreatedBy} 	&$96$
&\texttt{(\underline{Bread},CreatedBy,Flour)}, \texttt{(\underline{Cheese},CreatedBy,Milk)} \\
& & & \\ [-2ex]
\hline
& & & \\ [-2ex]
\multirow{3}{*}{WebChild}
&\texttt{Smaller}, \texttt{Better}, & \multirow{3}{*}{$38576$}
&\texttt{(\underline{Motorcycle},Smaller,\underline{Car})}, \texttt{(\underline{Apple},Better,VitaminPill)},\\
&\texttt{Slower}, \texttt{Bigger},  &
&\texttt{(\underline{Train},Slower,\underline{Plane})}, \texttt{(Watermelon,Bigger,\underline{Orange})}, \\
&\texttt{Taller},	$\dots$      &
&\texttt{(\underline{Giraffe},Taller,Rhino)}, \bluett{\texttt{(\underline{Skating},Faster,\underline{Walking}})}  
\\[-2ex] \\ 
\hline
\end{tabular}}
\footnotesize
\caption{ The relationships in different knowledge bases used for generating questions.
The `\#Facts' column shows the number of facts which are related to the visual concepts described in Section~\ref{sec:imgvc}.
The `Examples' column gives some examples of extracted facts, in which the visual concept is underlined.
}
\label{tab:predicates}
}
\vspace{-5pt}
\end{table*}

The knowledge about each visual concept is extracted from a range of existing structured knowledge bases,
including DBpedia~\cite{auer2007dbpedia}, ConceptNet~\cite{liu2004conceptnet} and WebChild~\cite{tandon2014acquiring}.

\noindent{\bf $-$ DBpedia:}
The structured information stored in DBpedia is extracted from Wikipedia by crowd-sourcing.
In this KB, concepts are linked to their categories and super-categories based on the SKOS Vocabulary\footnote{http://www.w3.org/2004/02/skos/}.
In this work, the categories and super-categories of all aforementioned visual concepts are extracted transitively.

\noindent{\bf $-$ ConceptNet:}
This KB is made up of several commonsense relations, such as \texttt{UsedFor}, \texttt{CreatedBy} and \texttt{IsA}.
Much of the knowledge is automatically generated from the sentences of
the Open Mind Common Sense (OMCS) project\footnote{http://web.media.mit.edu/{\textasciitilde}push/Kurzweil.html}.
We adopt $11$ common relationships in ConceptNet to generate questions and answers.

\noindent{\bf $-$ WebChild:}
The work in \cite{tandon2014acquiring} considered a form of commonsense knowledge being overlooked by most of existing KBs,
which involves comparative relations such as \texttt{Faster}, \texttt{Bigger} and \texttt{Heavier}.
In \cite{tandon2014acquiring}, this form of information is extracted automatically from the Web.

The relationships which we extract from each KB and the corresponding number of facts can be found in Table~\ref{tab:predicates}.
All the aforementioned structured information are stored in the form of RDF triples and can be accessed using Sparql queries.

\subsection{Question Collection}
\label{sec:quescollect}

In this work, we focus on collecting visual questions which need to be answered with the help of supporting-facts.
To this end, we designed a specialized system,
in which the procedure of asking questions is conducted in the following steps:

\begin{enumerate}[label={\arabic*)}]
\item {\em Selecting Concept:} Annotators are given an image and
a number of visual concepts (object, scene and action).
They need to choose one of the visual concepts which is related to this image.

\item {\em Selecting Fact:}
Once a visual concept is selected,
the associated facts are demonstrated in the form of sentences with the two entities underlined.
For example, the fact \texttt{(Train,Slower,Plane)} is expressed as `\underline{Train} is slower than \underline{plane}'.
Annotators should select a correct and relevant fact by themselves.

\item {\em Asking Question and Giving Answer:}
The Annotators are required to ask a question, answering which needs the information from
both of the image and the selected fact.
By doing so, the selected fact becomes the supporting-fact of the asked question.
The answer is limited to the two concepts in the supporting-fact.
In other words, the source of the answer can be either the visual concept grounded in the questioned image (underlined in Table~\ref{tab:predicates}) or the concept found on the KB side.

\end{enumerate}

\begin{table*}
	\label{tab:datasets}
	\centering
	\footnotesize
	\resizebox{0.85\linewidth}{!}{
		\rowcolors{3}{gray!15}{}
		\begin{tabular}{lccccccc}
			\hline
			~		&  Number of & Number of	& Num. question 	& Average quest.	& Average ans.&Knowledge&Supporting-	 \\
			Dataset	&  images	& questions	& categories		& length		& length&Bases&Facts	 \\ \hline
			DAQUAR \cite{malinowski2014multi}  &   1,449  &  12,468  &   4    &  11.5  &  1.2&-&-   \\
			COCO-QA \cite{ren2015image}  &   117,684  &  117,684  &    4    &  8.6  &  1.0 &-&-  \\
			VQA-real \cite{antol2015vqa}  &    204,721  &  614,163  &    20+    &  6.2  &  1.1 &-&-  \\
			Visual Genome \cite{krishnavisualgenome}  &    108,000  &  1,445,322  &    7   &  5.7  &  1.8  &-&- \\
			Visual7W \cite{zhu2015visual7w}  &   47,300  &  327,939  &    7    &  6.9  &  1.1 &-&-  \\
			Visual Madlibs \cite{Yu_2015_ICCV}  &    10,738  &  360,001  &    12    &  6.9  &  2.0 &-&- \\
			VQA-abstract \cite{antol2015vqa}  &    50,000  &  150,000  &    20+    & 6.2   &  1.1  &-&- \\
			VQA-balanced \cite{zhang2015yin}  &    15,623  &  33,379  &    1    &  6.2  &  1.0 &-&-  \\
			KB-VQA \cite{wang2015explicit}  &    700  &  2,402  &   23    & 6.8   & 2.0 &1&-  \\
			Ours (FVQA)  &    2,190  &  5,826   &  32    &  9.5  & 1.2&3& \checkmark  \\
			\hline
	\end{tabular}}
	\caption{Major datasets for VQA and their main characteristics.}
	\label{tab:datalen}
	\vspace{-3pt}
\end{table*}

\begin{table}[tpb!]
	\centering
	\resizebox{\linewidth}{!}
	{
		\begin{tabular}{llccc}
			\hline
			& & & &\\ [-2ex]
			Criterion &Category &Train &Test &Total \\
			& & & &\\ [-2ex]
			\hline
			& & & &\\ [-2ex]
			\multirow{3}{*}{%
				\hspace{-0.4cm}
				\begin{tabular}{l}
					Key Visual  \\ Concept (\texttt{$T_{\mbox{KVC}}$})    
				\end{tabular}	
			}
			& \texttt{Object}  &$2661.2\pm66.0$ &$2621.8\pm66.0$ &$5283$	\\
			& \texttt{Scene}   &$251.2\pm26.2$ &$260.8\pm26.2$ &$512$	\\
			& \texttt{Action}  &$14.8\pm2.7$ &$16.2\pm2.7$ &$31$\\
			& & & &\\ [-2ex]
			\hline
			& & & &\\ [-2ex]
			\multirow{2}{*}{%
				\hspace{-0.4cm}
				\begin{tabular}{l}
					Answer Source  \\ (\texttt{$T_{\mbox{AS}}$})    
				\end{tabular}
			}
			& \texttt{Image}   &$2437.4\pm63.4$ &$2393.6\pm63.4$ &$4831$		\\
			& \texttt{KB}      &$489.8\pm27.9$ &$505.2\pm27.9$ &$995$		\\
			& & & &\\ [-2ex]
			\hline
			& & & &\\ [-2ex]
			\multirow{3}{*}{%
				\hspace{-0.4cm}
				\begin{tabular}{l}
					KB Source  \\ (\texttt{$T_{\mbox{KS}}$})    
				\end{tabular}
			}
			& \texttt{DBpedia}    &$403.2\pm12.7$ &$413.8\pm12.7$ &$817$		\\
			& \texttt{ConceptNet} &$2348.8\pm71.6$ &$2303.2\pm71.6$ &$4652$\\
			& \texttt{Webchild}   &$175.2\pm9.5$ &$181.8\pm9.5$ &$357$\\
			& & & &\\ [-2ex]
			\hline
			& & & &\\ [-2ex]
			Total
			&            &$2927.2\pm69.5$ &$2898.8\pm69.5$ &$5826$\\
			& & & &\\ [-2ex]
			\hline
	\end{tabular}}
	\caption{
		The classification of questions according to Key Visual Concept, 
		KB Source and Answer Source.
		The number of training/test questions in each category is also demonstrated. 
		The error bars are produced by $5$ different splits.
	}
	\label{tab:trntst}
\end{table}

\subsection{Data Statistics}
\label{sec:Statistics}
\bluetext{
\bluett{ 
In total, $5826$ questions (corresponding to $4216$ unique facts) are collected collaboratively by $38$ volunteers. 
The average annotation time for each question is around $1$ minute. }
In order to report significant statistics, we create 5 random splits of the dataset.
In each split, we have $1100$ training images and $1090$ test images. Each split provides roughly $2927$ and $2899$ questions\footnote{As each image contains a different number of questions, each split may contain different number of questions for training and test. Here we only report the average numbers. The error bars in the Table~\ref{tab:trntst} show the differences.} for training and test respectively. 

\bluetext{
	Table~\ref{tab:datalen} shows summary statistics of the dataset, such as the number of question categories, average question/answer length \etc. We have $32$ question types in total (see Section \ref{subsec:QQ_Maping} for more details). Compared to VQA-real \cite{antol2015vqa} and Visual Genome \cite{krishnavisualgenome}, our FVQA dataset provides longer questions, with average length $9.5$ words.}

These questions can be categorized according to the following criteria:

\noindent \bluett{\bf $-$ Key Visual Concept}: 
\bluett{
The visual concept that appears in the supporting-fact of the given question, 
which is selected in Step-$1$ of the question collection process. 
The type of Key Visual Concept is denoted by \texttt{$T_{\mbox{KVC}}$}, whose value can be 
\texttt{Object}, \texttt{Scene} or \texttt{Action} (see Section~\ref{sec:imgvc}).}

\noindent \bluett{\bf $-$ Key Relationship}:
\bluett{
The relationship in the supporting-fact, whose type is denoted as \texttt{$T_{\mbox{REL}}$}.
As shown in Table~\ref{tab:predicates}, there are $13$ different values that can be assigned to \texttt{$T_{\mbox{REL}}$}\footnote{\bluettt{For simplicity, we consider all the comparative relationships in WebChild as one type.}}.}

\noindent \bluett{\bf $-$ KB Source} \bluett{(refer to as \texttt{$T_{\mbox{KS}}$}):
The external KB where the supporting-fact is stored. 
In this work, the value of \texttt{$T_{\mbox{KS}}$} can be 
\texttt{DBpedia}, \texttt{ConceptNet} or \texttt{Webchild}. }

\noindent \bluett{\bf $-$ Answer Source} \bluett{(refer to as \texttt{$T_{\mbox{AS}}$}):
As shown in Step3 of Section~\ref{sec:quescollect}, 
\texttt{$T_{\mbox{AS}}$} $=$ \texttt{Image} if answer is the key visual concept, 
and \texttt{$T_{\mbox{AS}}$} $=$ \texttt{KB} otherwise.}

Table~\ref{tab:trntst} shows the number of training/test questions falling into different categories of 
\texttt{$T_{\mbox{KVC}}$}, \texttt{$T_{\mbox{KS}}$}, \texttt{$T_{\mbox{AS}}$}.
We can see that most of the questions are related to the objects in images and most of the answers are from 
\texttt{Image}.
As for knowledge bases, $80\%$ of the collected questions rely on the supporting-facts from ConceptNet.
Answering $14\%$ and $6\%$ questions depends on the knowledge from DBpedia and Webchild respectively.

\begin{figure}[tbp!]
\begin{center}
	\includegraphics[width=0.99\columnwidth]{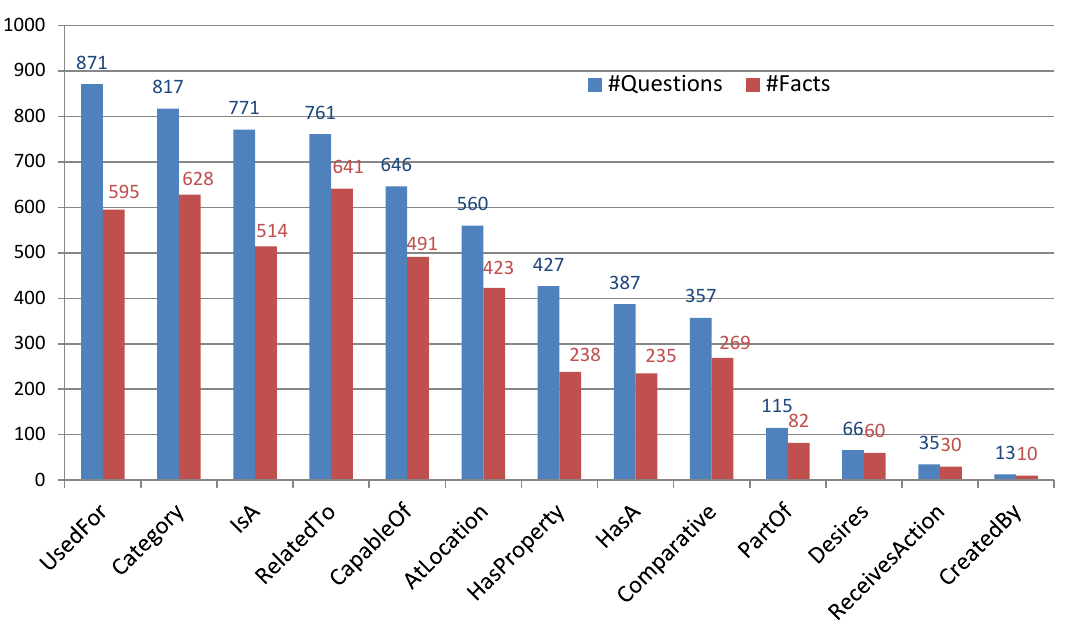}
\end{center}
\vspace{-5pt}
\caption{The distributions of the collected $5826$ questions and the corresponding $4216$ facts over different relationships.
The top five relationships are \texttt{UsedFor}, \texttt{Category}, \texttt{IsA}, \texttt{RelatedTo} and \texttt{CapableOf}. There are fewer supporting-facts than questions because one `fact' can correspond to multiple `questions'.}
\label{fig:predicates}
\end{figure}

The distributions of collected questions and facts over the $13$ types of key relationships (\texttt{$T_{\mbox{REL}}$}) are shown in Figure~\ref{fig:predicates}. 
We can see that the questions and facts are evenly distributed over the relationships of
\texttt{Category}, \texttt{UsedFor}, \texttt{IsA}, \texttt{RelatedTo}, \texttt{CapableOf},
\texttt{AtLocation}, \texttt{HasProperty} and \texttt{HasA},
although these relationships differ significantly in the total numbers of extracted facts (see Table~\ref{tab:predicates}).

\subsection{Human Study of Common-sense Knowledge}
In order to verify whether our collected questions require common-sense knowledge and whether the supporting-facts are helpful for answering the knowledge required questions, we conducted two human studies by asking subjects:
\begin{enumerate}
 \item Whether or not the given question requires external common-sense knowledge to answer, and If `yes'
 \item Whether or not the given supporting-fact provides the common-sense knowledge to answer the question.
\end{enumerate}
The above study is repeated by 3 human subjects independently.
\bluettt{
We found that $97.6\%$ of collected questions are voted as `require common-sense knowledge' by at least 2 subjects. For these knowledge-requiring questions, more than $99\%$ of the supporting-facts provided in our dataset 
are considered valuable for answering them.}

\subsection{Comparison}

The most significant difference between the proposed dataset and existing VQA datasets
is on the provision of supporting-facts.
A large portion of visual questions require not only the information from the image itself,
but also the often overlooked but critical commonsense knowledge external to the image.
\bluetext{It is shown in \cite{antol2015vqa} that 3 or more subjects agreed that
$47.43\%$ questions in the VQA dataset require commonsense reasoning to answer ($18.14\%$:  6 or more subjects).
However, such external knowledge is not provided in all the existing VQA datasets.
To the best knowledge of us, this is the first VQA dataset providing supporting-facts.}

In this dataset, the supporting-facts which are necessary for answering the corresponding visual questions
are obtained from several large-scale structured knowledge bases.
This dataset enables the development of approaches which utilize the information from both the image and the external knowledge bases. \bluetext{Different from \cite{wang2015explicit} that only applied a single manually annotated knowledge source, we use two additional automatically extracted knowledge bases, which enable us to answer more general questions.}

In a similar manner as ours, the Facebook bAbI~\cite{weston2015towards} dataset also provides supporting-facts for pure textual questions.
But the problem posed in this work is more complex than that in Facebook bAbI, as the information need to be extracted from both image and external commonsense knowledge bases.

\bluett{Another feature of the proposed dataset is that the answers are restricted to the concepts from image and knowledge bases,
so `Yes'/`No' questions are excluded.
In the VQA dataset~\cite{antol2015vqa}, $38\%$ questions can be answered using `Yes' or `No'.
Although `Yes'/`No' questions can also require a challenging reasoning process (see \cite{andreas2015deep,johnson2016clevr}), 
{\bluettt{they may not be a good measure of models' reasoning abilities.
A random guess process can still achieve an approximately $50\%$ accuracy on a balanced `Yes'/`No' QA dataset, but it does not perform any reasoning on the $50\%$ correctly answered questions.}}
To avoid misleading results, we simply exclude `Yes'/`No' questions.}

\section{Approach}
\label{method}
As shown in Section~\ref{dataset}, all the information extracted from images and KBs are stored as a graph of interlinked RDF triples.
State-of-the-art RNN approaches~\cite{gao2015you,malinowski2015ask,zhu2015visual7w,Chen2015ABC,Jiang2015compositional,andreas2015deep,yang2015stacked} directly learn the mapping between questions and answers,
which, however, do not scale well to the diversity of answers and cannot provide the key information that the reasoning is based on.
In contrast, we propose to learn the mapping between questions and a set of KB-queries,
such that there is no limitation to the vocabulary size of the answers (\ie, the answer to a test question does not have to be observed ahead in the training set)
and the supporting-facts used for reasoning can be provided.

\begin{figure*}[tbp!]
\begin{center}
	\includegraphics[width=1.99\columnwidth]{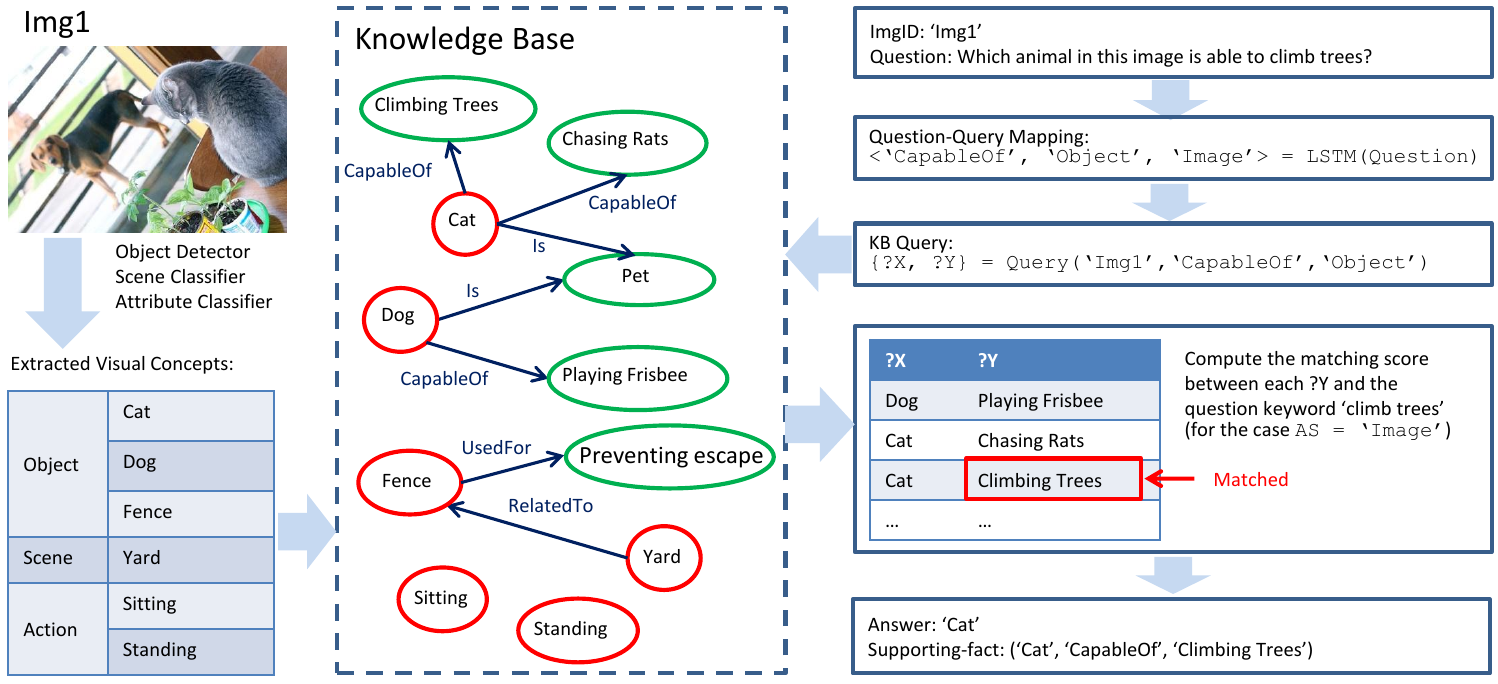}
\end{center}
\caption{An example of the reasoning process of the proposed VQA approach.
The visual concepts (objects, scene, attributes) of the input image are extracted using trained models, which are further linked to the corresponding semantic entities in the knowledge base.
The input question is firstly mapped to one of the query types using the LSTM model shown in Section~\ref{subsec:QQ_Maping}.
The types of key relationships, key visual concept and answer source can be determined accordingly.
A specific query (see Section~\ref{sec:querying}) is then performed to find all facts meeting the search conditions in KB.
These facts are further matched to the keywords extracted from the question sentence.
The fact with the highest matching score is selected and the answer is also obtained accordingly.}
\label{fig:approach}
\end{figure*}

\subsection{Constructing a Unified KB}
\bluett{
The primary step of our approach is to construct a unified KB, 
which links the visual concepts extracted from each image to the corresponding concepts in multiple KBs.
A visual concept $X$ of type $T$ ($T=$ \texttt{Object}, \texttt{Sence} or \texttt{Action}) 
extracted from the image with id $I$ is stored in two triples 
\texttt{($X$,Grounded,$I$)} and \texttt{($X$,VC-Type,$T$)}.
Concepts in multiple KBs with the same meaning as $X$ are directly linked to $X$.
By doing so, the rich external knowledge about $X$ is linked to the images that $X$ is grounded in.
} 
 
\subsection{Question-Query Mapping}
\label{subsec:QQ_Maping}
In our approach, three characteristics of visual questions are firstly predicted by trained LSTM models, \ie, 
key visual concepts (\texttt{$T_{\mbox{KVC}}$}), key relationships (\texttt{$T_{\mbox{REL}}$}) and answer sources (\texttt{$T_{\mbox{AS}}$}). 
In the training data, these characteristics of a question can be obtained through the annotated supporting-fact and the given answer,
and we have collected $32$ different combinations of $\langle$\texttt{\texttt{$T_{\mbox{KVC}}$},\texttt{$T_{\mbox{REL}}$},\texttt{$T_{\mbox{AS}}$}}$\rangle$ in the proposed dataset (see Appendix). Since both question and query are sequences, the question-query mapping problem can be treated as a sequence-to-sequence problem \cite{sutskever2014sequence}, which can be solved by Recurrent Neural Network (RNN) \cite{mikolov2010recurrent}. While in this work, we consider each distinct combination of the three characteristics as a query type and learn a $32$-class classifier using LSTM models\cite{hochreiter1997long}, in order to identify the above three properties of an input question and perform a specific query.}

The LSTM is a memory cell encoding knowledge at every time step for what inputs have been observed up to this step. We follow the model used in \cite{qi2015caption}. 
The LSTM model for the question to query type mapping is trained in an unrolled form. More formally, the LSTM takes the sequence of words in the given question $Q=(Q_0,...,Q_L)$, where $Q_0$ is a special start word. Each word has been represented as a one-hot vector $S_t$. At time step $t=0$, we set $x_0=W_{es}S_0$ and $h_{initial}=\vec{0}$, where $W_{es}$ is the learnable word embedding weights. From $t=1$ to $t=L$, we set $x_t=W_{es}S_t$ and the input hidden state $h_{t-1}$ is given by the previous step. The cost function is
\begin{equation}
 \mathcal{C}=-\frac{1}{N}\sum_{i=1}^N\log p(T^{(i)})+\lambda_\ModParms\cdot||\ModParms||_2^2
\end{equation}
where $N$ is the number of training examples. $T^{(i)}$ is the ground truth query types of the $i$-th training question. $\log p(T^{(i)})$ is the log-probability distribution over all candidate query types that is computed by the last LSTM cell, given the previous hidden state and the last word of question. $\ModParms$ represents model parameters, $\lambda_\ModParms\cdot||\ModParms||_2^2$ is a regularization term.

During the test phase, the question's words sequence is fed into the trained LSTM to produce the probability distribution $p$ over all query types, via equations (1) to (7).

In Figure~\ref{fig:approach}, the query type of the input question 
{\em `Which animal in this image is able to climb trees?'}
is classified by the LSTM classifier as $\langle$\texttt{\texttt{$T_{\mbox{REL}}$},\texttt{$T_{\mbox{KVC}}$},\texttt{$T_{\mbox{AS}}$}}$\rangle$$=$
$\langle$\texttt{CapableOf,Object,Image}$\rangle$.

\subsection{Querying the KB}
\label{sec:querying}

\bluett{
Retrieving the correct supporting-fact is the key to answering a visual question in our proposed dataset. 
To this end, KB queries are constructed to search for a number of candidate supporting-facts.
To be specific, 
given a question's key relationship (\texttt{$T_{\mbox{REL}}$}) and key visual concept (\texttt{$T_{\mbox{KVC}}$}) as inputs,
a KB-query $\{ \mbox{\texttt{?X,?Y}} \} = \mbox{\texttt{Query($I$,$T_{\mbox{REL}}$,$T_{\mbox{KVC}}$)}}$ is constructed as follows:
\begin{align}
\mbox{Find \texttt{?X}, \texttt{?Y}, subject to }
\{&\mbox{\texttt{(?X,Grounded,$I$)} and} \notag \\
&\mbox{\texttt{(?X,VC-Type,$T_{\mbox{KVC}}$)}}\,\, \mbox{and} \notag \\
&\mbox{\texttt{(?X,$T_{\mbox{REL}}$,?Y)}} \}, \quad\,\,\,\, \notag
\end{align}
where $I$ denotes the id of questioned image; 
\texttt{Grounded} is a relationship representing that a specific visual concept is {\em grounded} in a specific image;
\texttt{VC-Type} is another relationship used to describe the type of a visual concept;
\texttt{?X} and \texttt{?Y} are two variables to be searched and returned.
There are three triple templates to be matched:
\texttt{(?X,Grounded,$I$)} searches all visual concepts \texttt{?X} that are grounded in image $I$;
\texttt{(?X,VC-Type,$T_{\mbox{KVC}}$)} restricts that the type of \texttt{?X} should be \texttt{$T_{\mbox{KVC}}$};
\texttt{(?X,$T_{\mbox{REL}}$,?Y)} restricts that \texttt{?X} should be linked to at least one concept  
\texttt{?X} via relationship \texttt{$T_{\mbox{REL}}$}.
The above query will search for sets of triples that match these three triple templates,
and return a list of $\{$\texttt{?X,?Y}$\}$ pairs residing in the associated slots of matched sets.
Correspondingly, we also obtain a list of candidate supporting-facts \texttt{(?X,$T_{\mbox{REL}}$,?Y)}.
Note that the query is performed over the entire KB that stores all facts about visual concepts,
rather than over the $4216$ supporting-facts that have been used in the collected questions.
As aforementioned, all comparative relationships are considered as one relation type ($T_{\mbox{REL}}$). 
In this case, the returned query results will be further filtered based on the comparative words shown in questions.}

\bluett{
In the reasoning process shown in Figure~\ref{fig:approach}, for all objects in Image \texttt{Img1}, the query \texttt{Query(Img1,CapableOf,Object)}
searches  for the things they are capable of doing.
The objects `Dog' and `Cat' in this image have links 
to concepts in the constructed KB via relationship \texttt{CapableOf}.
Accordingly, a list of $\{$\texttt{?X,?Y}$\}$ pairs are returned by this query，
including \texttt{\{Dog,Playing Frisbee\}, \{Cat,Chasing Rats\} and \{Cat,Climbing Trees\}}.
More examples of KB-queries and their outputs can be found in Figure~\ref{results_examples}.
}

\subsection{Answering}
\label{sec:answering}

\bluett{
As shown in the Step-$3$ of Section~\ref{sec:quescollect}, 
the answer to a visual question in our collected dataset can be either the visual concept
in the supporting-fact (\ie, \texttt{?X}), or the other concept on the KB side 
(\ie, \texttt{?Y}).
In the test phase, whether the answer is \texttt{?X} or \texttt{?Y} is determined by the value of
answer source \texttt{$T_{\mbox{AS}}$} predicted by the LSTM classifier in Section~\ref{subsec:QQ_Maping}: 
the answer is \texttt{?X} if \texttt{$T_{\mbox{AS}}$ = Image}, and \texttt{?Y} if \texttt{$T_{\mbox{AS}}$ = KB}.
The last issue before arriving at the answer is how to select the most relevant supporting-fact from candidates.
In this work, the supporting-fact is selected by matching between the given question and 
concepts \texttt{?Y} (or \texttt{?X}) in candidate facts if answer is \texttt{?X} (or \texttt{?Y}), 
as follows:}

\bluett{
1) {\texttt{$T_{\mbox{AS}}$ = Image}}:
A list of high frequency words (such as `what', `which', `a', `the') 
is established by counting in the training examples.
The keywords of a question are then extracted by removing these high frequency words.
A matching score $s$ is computed between the question keywords and 
the concept \texttt{?Y} in each candidate supporting-fact \texttt{(?X,$T_{\mbox{REL}}$,?Y)},
in order to measure the relevance between the fact and question. 
In this work, the matching score is the simple Jaccard similarity between the normalized word sets of \texttt{?Y}
(refer to as $\mathcal{W}_{\mathrm{Y}}$) and question keywords (refer to as $\mathcal{W}_{\mathrm{Q}}$):
\begin{align}
{s}(\mathcal{W}_{\mathrm{Y}}, \mathcal{W}_{\mathrm{Q}}) 
		= \frac{|\mathcal{W}_{\mathrm{Y}} \cap \mathcal{W}_{\mathrm{Q}}|}{ |\mathcal{W}_{\mathrm{Y}} \cup \mathcal{W}_{\mathrm{Q}}|}.
\end{align}
The candidate fact corresponding to the highest-scored \texttt{?Y} is selected as the supporting-fact,
and the associated visual concept \texttt{?X} is considered as the answer.
Note that the matching score can be computed using word embedding models such as Word2Vec~\cite{mikolov2013efficient}, which we leave for future work.
In the example of Figure~\ref{fig:approach}, all \texttt{?Y}s (\ie \texttt{Playing Frisbee}, \texttt{Chasing Rats} and \texttt{Climbing Trees}) are matched to the question keywords \texttt{climb trees} and
the fact \texttt{(Cat,CapableOf,Climbing Trees)} has achieved the highest score, so the answer is \texttt{Cat}.
}

\bluett{
2) {\texttt{$T_{\mbox{AS}}$ = KB}}:
In this case, 
we need to find out which visual concept (\texttt{?X}) is the most related to the input question.
If \texttt{$T_{\mbox{KVC}}$} $=$ \texttt{Scene} or \texttt{Action},
the visual concept \texttt{?X} with the highest probability 
is selected and the corresponding concept \texttt{?Y} is considered as the answer.
The probabilities of scene or action concepts are obtained 
from the softmax layer of the visual models shown in Section~\ref{sec:imgvc}.
If \texttt{$T_{\mbox{KVC}}$} $=$ \texttt{Object},
the visual concept \texttt{?X} is selected based on the question keywords describing location
(such as \texttt{top}, \texttt{bottom}, \texttt{left}, \texttt{right} or \texttt{center})
or size (such as \texttt{small} and \texttt{large}).
{ Note that one visual concept \texttt{?X} may correspond to multiple concepts \texttt{?Y}, 
\ie, multiple answers.
These answers are ordered according to their frequency in the training data:
the most frequent answer appears first.}
}

\section{Experiments}
\label{expe}
In this section, we first evaluate the question to KB-query mapping performance of our models. As a key component of our models, its performance impacts the final visual question answering (VQA) accuracy. We then report the performance of several baseline models, comparing with the proposed method. Different from all the baseline models, our method is able to perform explicit reasoning, that is, we can select the supporting-fact from knowledge bases that leads to the answer. We also report the supporting-fact selection accuracy.

\subsection{Question-Query Mapping}
Table~\ref{question_query_map} reports the accuracy of our proposed Question-Query mapping (QQmaping) approach in Sec \ref{subsec:QQ_Maping}. The mapping model is trained on the \KBName training splits and tested on the respective testing splits. To train the model, we use the Stochastic Gradient Descent (SGD) with $100$ Question-Query pairs as a mini-batch. Both the word embedding size and the LSTM memory cell size are $128$. The learning rate is set to $0.001$ and clip gradient is $10$. The dropout rate is set to $0.5$. It converged after $50$ epochs of training. We also provide the results for different KB Sources of supporting facts. Questions asked based on facts from WebChild are much easier to be mapped than questions based on other two KBs. This is mainly because facts in WebChild are related to the `comparative' relationship, such as `car is faster than bike', which further lead to user-generated questions are more repeated in similar formats. For example, many questions are formulated as `Which object in the image is more $<$\textit{a comparative adj}$>$ ?'. Our Top-3 overall accuracy achieves $82.42\pm0.56$.

\begin{table}[t!]
\centering
\resizebox{0.63\linewidth}{!}{
\begin{tabular}{lcc}
\hline
\multicolumn{1}{c}{\textbf{KB Source}} & \multicolumn{2}{c}{Question-Query Mapping Acc. $\pm$ Std (\%)} \\ \cline{2-3}
\multicolumn{1}{c}{\textbf{}}      & \textbf{Top-1}            & \textbf{Top-3}           \\ \hline
DBpedia                                & $61.73\pm1.83$                    & $85.56\pm2.14$	                    \\
ConceptNet                             & $64.10\pm1.10$                    & $80.85\pm0.87$                    \\
WebChild                               & $83.15\pm3.03$                    & $95.24\pm1.23$                    \\ \hline
Overall                                & $64.94\pm1.08$                    & $82.42\pm0.56$	                   \\ \hline
\end{tabular}
}
\caption{Question-Query mapping accuracy for different KB sources on the \KBName testing splits. Top-$1$ and Top-$3$ results are reported.}
\label{question_query_map}
\end{table}

\subsection{\KBName Experiments}

Our \KBName tasks are formulated as an open-ended answer generation problem, which requires models to predict open-ended outputs that may not appear in training data. To measure the accuracy, we simply calculate the proportion of correctly answered test questions. A predicted answer is determined as correct if and only if its string matches the corresponding ground-truth answer (all the answers have been normalized by the python INFLECT package to eliminate the singular-plurals differences \etc). We report the top-$1$, $3$ and $10$ accuracy of the evaluated methods. { Note that our proposed models may produce less than $10$ candidate answers in some cases where the number of returned supporting facts is less than $10$, which makes the top-$10$ performance of our models not as strong as on the top-$1$ and top-$3$ settings.}

Additionally, we also report the results based on Wu-Palmer Similarity (WUPS)~\cite{wu1994verbs}. WUPS calculates the similarity between two words based on their common subsequence in the taxonomy tree. If the similarity is greater than a prescribed threshold then the predicted answer is considered as correct. In this paper, we report WUPS results at thresholds $0.9$ and $0.0$. All the reported results are averaged on the $5$ test splits (standard deviation is also provided). {It should be noted that WUPS@$0.0$ is calculated with a very small threshold, which 
	offers a very loose similarity measure.}

\begin{table*}[h!]
\centering
\resizebox{.657\linewidth}{!}{
\begin{tabular}{lccc}
\hline
\multicolumn{1}{c}{\multirow{2}{*}{\textbf{Method}}} & \multicolumn{3}{c}{Overall Acc. $\pm$ Std (\%)} \\ \cline{2-4}
\multicolumn{1}{c}{}                         & \textbf{Top-1}   & \textbf{Top-3}   & \textbf{Top-10}  \\ \hline
SVM-Question                                 & $10.37\pm0.80$   & $20.72\pm0.58$   & $34.63\pm1.19$   \\
SVM-Image                                    & $18.41\pm1.07$   & $32.42\pm1.06$   & $47.53\pm1.02$   \\
SVM-Question+Image                           & $18.89\pm0.91$   & $32.78\pm0.90$   & $48.13\pm0.73$   \\
LSTM-Question                                & $10.45\pm0.57$   & $19.02\pm0.74$   & $31.64\pm0.93$   \\
LSTM-Image                                   & $20.55\pm0.81$   & $36.01\pm1.45$   & $55.74\pm2.28$  \\
{LSTM-Question+Image}                          & {$ 22.97\pm0.64$}   & {$36.76\pm1.22$}   & {$54.19\pm2.45$}   \\ 
{LSTM-Question+Image+Pre-VQA}                  & {$24.98\pm0.60$}   & {$40.40\pm1.05$}   & {$57.27\pm1.29$}   \\ 
{Hie-Question+Image}                           & {$33.70\pm1.18$}	& {$50.00\pm0.78$}   & {$64.08\pm0.57$} \\
{Hie-Question+Image+Pre-VQA}                   & {$43.14\pm0.61$}	& {$59.44\pm0.34$}   & {$\mathbf{72.20\pm0.39}$}  \\
\hline \hline
\cellcolor[rgb]{0.8,0.8,0.8}{Ours, gt-QQmaping\textsuperscript{$\ddagger$}}                               & \cellcolor[rgb]{0.8,0.8,0.8}{$63.63\pm0.73$}   & \cellcolor[rgb]{0.8,0.8,0.8}{$71.30\pm0.78$}   & \cellcolor[rgb]{0.8,0.8,0.8}{$72.55\pm0.79$}   \\
Ours, top-1-QQmaping                            & $52.56\pm1.03$   & $59.72\pm0.82$   & $60.58\pm0.86$   \\
Ours, top-3-QQmaping                            & $\mathbf{56.91\pm0.99}$   & $\mathbf{64.65\pm1.05}$   & ${65.54\pm1.06}$   
\\ \hline \hline
{Ensemble}                                     & {$58.76\pm0.92$} &- &- \\
\hline\hline
Human                                        & $77.99\pm0.75$
                                             &- &- \\ \hline
\end{tabular}}
\caption{Overall accuracy on our \KBName testing splits for different methods based on string matching. 
	Best single-model results are shown in bold font.
	$\ddagger$~indicates that ground truth Question-Query mappings are used, which (in \colorbox[rgb]{0.8,0.8,0.8}{gray}) will not participate in rankings.}
\label{over-acc}
\end{table*}

\begin{table*}[t!]
\centering
\resizebox{.66\linewidth}{!}{
\begin{tabular}{lccc}
\hline
\multicolumn{1}{c}{\multirow{2}{*}{\textbf{Method}}} & \multicolumn{3}{c}{WUPS@0.9. $\pm$ Std (\%)} \\ \cline{2-4}
\multicolumn{1}{c}{}                         & \textbf{Top-1}   & \textbf{Top-3}   & \textbf{Top-10}  \\ \hline
SVM-Question                                 & $17.06\pm1.09$   & $29.43\pm0.62$   & $44.73\pm1.07$   \\
SVM-Image                                    & $24.73\pm1.29$   & $40.95\pm1.34$   & $56.33\pm0.63$   \\
SVM-Question+Image                           & $25.30\pm1.09$   & $41.37\pm1.19$   & $56.78\pm0.64$   \\
LSTM-Question                                & $15.82\pm0.57$   & $26.45\pm0.61$   & $40.99\pm1.02$   \\
LSTM-Image                                   & $26.78\pm1.02$   & $44.00\pm1.61$   & $62.86\pm2.23$  \\
{LSTM-Question+Image}                          & {$29.08\pm0.91$}   & {$44.36\pm1.29$}   & {$61.71\pm2.82$}   \\ 
{LSTM-Question+Image+Pre-VQA}                  & {$31.96\pm0.65$}   & {$48.55\pm0.97$}   & {$64.73\pm1.38$}   \\ 
{Hie-Question+Image}                           & {$39.75\pm0.78$}   & {$56.48\pm0.68$}   & {$69.78\pm0.51$}   \\ 
{Hie-Question+Image+Pre-VQA}					 & {$48.93\pm0.71$}   & {$64.75\pm0.44$}   & {$\mathbf{76.73\pm0.38}$}   \\ 
\hline \hline
\cellcolor[rgb]{0.8,0.8,0.8}{Ours, gt-QQmaping\textsuperscript{$\ddagger$}}                               & \cellcolor[rgb]{0.8,0.8,0.8}{$65.51\pm0.82$}   & \cellcolor[rgb]{0.8,0.8,0.8}{$72.37\pm0.89$}   & \cellcolor[rgb]{0.8,0.8,0.8}{$73.55\pm0.87$}   \\
Ours, top-1-QQmaping                            & $54.79\pm0.91$   & $61.41\pm0.71$   & $62.22\pm0.70$   \\
Ours, top-3-QQmaping                            & $\mathbf{59.67\pm0.90}$   & $\mathbf{66.89\pm1.01}$   & ${67.77\pm1.04}$   \\ \hline\hline
Human & $82.47\pm0.71$
        &-&- \\ \hline
\end{tabular}}
\caption{WUPS@0.9 on our \KBName testing splits for different methods. 
	Best single-model results are shown in bold font.
	$\ddagger$~indicates that ground truth Question-Query mappings are used, which (in \colorbox[rgb]{0.8,0.8,0.8}{gray}) will not participate in rankings.}
\label{over-WUPS9}
\end{table*}

\begin{table*}[t!]
\centering
\resizebox{.66\linewidth}{!}{
\begin{tabular}{lccc}
\hline
\multicolumn{1}{c}{\multirow{2}{*}{\textbf{Method}}} & \multicolumn{3}{c}{WUPS@0.0. $\pm$ Std (\%)} \\ \cline{2-4}
\multicolumn{1}{c}{}                         & \textbf{Top-1}   & \textbf{Top-3}   & \textbf{Top-10}  \\ \hline
SVM-Question                                 & $56.88\pm2.57$   & $68.45\pm0.67$   & $76.93\pm0.75$   \\
SVM-Image                                    & $59.64\pm0.72$   & $73.30\pm0.96$   & $81.77\pm0.33$   \\
SVM-Question+Image                           & $59.97\pm0.63$   & $73.39\pm0.88$   & $81.93\pm0.39$   \\
LSTM-Question                                & $51.45\pm1.41$   & $65.65\pm0.66$   & $75.76\pm0.62$   \\
LSTM-Image                                   & $59.53\pm1.52$   & $73.83\pm0.50$   & $83.53\pm0.89$  \\
{LSTM-Question+Image}                          & {$61.86\pm0.81$}   & {$74.45\pm0.59$}   & {$83.54\pm0.94$}   \\
{LSTM-Question+Image+Pre-VQA}                  & {$63.42\pm0.35$}   & {$76.63\pm0.50$}   & {${84.94\pm0.57}$}   \\
{Hie-Question+Image}                           & {$66.11\pm0.62$}   & {$78.90\pm0.50$}   & {$86.32\pm0.55$}   \\ 
{Hie-Question+Image+Pre-VQA}					 & {$71.51\pm0.16$}   & {$\mathbf{82.71\pm0.39}$}   & {$\mathbf{89.01\pm0.50}$}   \\ 
\hline \hline
\cellcolor[rgb]{0.8,0.8,0.8}{Ours, gt-QQmaping\textsuperscript{$\ddagger$}}                               & \cellcolor[rgb]{0.8,0.8,0.8}{$73.98\pm0.77$}   & \cellcolor[rgb]{0.8,0.8,0.8}{$78.67\pm0.78$}   & \cellcolor[rgb]{0.8,0.8,0.8}{$79.98\pm0.79$}   \\
Ours, top-1-QQmaping                            & $64.96\pm0.70$   & $69.57\pm0.65$   & $70.64\pm0.59$   \\
Ours, top-3-QQmaping                            & $\mathbf{72.34\pm0.65}$   & ${77.52\pm0.70}$   & $78.69\pm0.63$   \\ \hline\hline
Human & $87.30\pm0.54$
      &- &-\\ \hline
\end{tabular}}
\caption{WUPS@0.0 on our \KBName testing splits for different methods. 
	Best single-model results are shown in bold font.
	$\ddagger$~indicates that ground truth Question-Query mappings are used, which (in \colorbox[rgb]{0.8,0.8,0.8}{gray}) will not participate in rankings.}
\label{over-WUPS0}
\end{table*}

We evaluate baseline models on the \KBName tasks in three sets of experiments: without images (Question), without questions (Image) and with both images and questions (Question+Image). Same as \cite{zhu2015visual7w}, in the experiments without images (or questions), we zero out the image (or question) features. We briefly describe the three models we used in the experiments:

\noindent{\textbf{SVM.}}
A Support Vector Machine (SVM) model that predicts the answer from a concatenation of image features and question features. For image features, we use the fc7 layer ($4096$-d) of the VggNet-16 model~\cite{simonyan2014very}. Questions are represented by $300$-d averaged word embeddings from a pre-trained word2vec model \cite{mikolov2013distributed}. We take the top-$500$ most frequent answers ($93.68\%$ of the training set answers) as the class labels. At test time, we select the top-$1$, $3$ and $10$ scored answer candidates. We use the LibSVM~\cite{chang2011libsvm} and the parameter $C$ is set to the default value $1$. { We found that tuning the the value of $C$ does not affect the performance significantly.}

\noindent{\textbf{LSTM.}}
\bluetext{We compare our system with an approach \cite{ren2015image} based on LSTM. The LSTM outputs at the last timestep are fed into a softmax layer to predict answers over a fixed space (top-$500$ most frequent answers). This is similar to the `LSTM+MLP' method proposed in \cite{antol2015vqa}. }Specifically, we use the fc7 layer ($4096$-d) of the pre-trained VggNet-16 model as the image features, and the LSTM is trained on our training splits. The LSTM layer contains $512$ memory cells in each unit. The learning rate is set to $0.001$ and clip gradient is $5$. The dropout rate is set to $0.5$. Same as SVM models, we select the top-$1$, $3$ and $10$ scored answer candidates at test time. { We additionally implement a model that is firstly pre-trained on the VQA dataset, and fine-tuned on the FVQA dataset, which is denoted as `LSTM-Question+Image+Pre-VQA'.}

{ \noindent{\textbf{State-of-the-art}.}
We evaluate the model in \cite{lu2016hierarchical} on FVQA with and without pre-training on the VQA dataset. The models are denoted as `Hie-Question+Image' and `Hie-Question+Image+Pre-VQA'. The hierarchical co-attention model of \cite{lu2016hierarchical} provides the state-of-the-art performance on the popular VQA dataset.}

{ \noindent{\textbf{Ensemble Model.}}
We combine two LSTM-based models
(which are `Hie-Question+Image+Pre-VQA' and `LSTM-Question+Image+Pre-VQA') with our proposed model (`Ours, top3-QQmaping') by selecting the answer with the maximum score over the three models. The scores of LSTM models are taken from the Softmax layer, and the score of our model is $0$ when it is not able to find any answer and $1$ otherwise).}

\noindent{\textbf{Human.}}
\bluetext{We also report the human performance. Questions in the test splits are given to $5$ human subjects and they are allowed to use any media (such as books, Wikipedias, Google \etc) to gather the information or knowledge to answer the questions. Human subjects are only allowed to provide one answer to one question, so there is no Top-$3$ and Top-$10$ evaluations for the human performance. Note that these $5$ subjects are never involved in the question collection procedure.}

\noindent{\textbf{Ours.}}
Our KB-query based model is introduced in Section \ref{method}. To verify the effectiveness of our method, we implement three variants. `{gt-QQmaping}' uses the ground truth question-query mapping,
while `{top-1-QQmaping}' and `{top-3-QQmaping}' use the top-$1$ and top-$3$ predicted question-query mapping (see Section \ref{subsec:QQ_Maping}), respectively.

\begin{table*}[t!]
	\centering
	\resizebox{\linewidth}{!}{
	\begin{tabular}{lccclccccccc}
		\hline
		\multicolumn{1}{c}{\multirow{3}{*}{\textbf{Method}}} & \multicolumn{11}{c}{KB-Source, Acc. $\pm$ Std (\%)}                                                                                        \\ \cline{2-12}
		\multicolumn{1}{c}{}                         & \multicolumn{3}{c}{\textbf{DBpedia}} & \multicolumn{1}{c}{} & \multicolumn{3}{c}{\textbf{ConceptNet}} &  & \multicolumn{3}{c}{\textbf{WebChild}} \\ \cline{2-4} \cline{6-8} \cline{10-12}
		\multicolumn{1}{c}{}                         & \textbf{Top-1}   & \textbf{Top-3}   & \textbf{Top-10}  &                      & \textbf{Top-1}    & \textbf{Top-3}    & \textbf{Top-10}   &  & \textbf{Top-1}   & \textbf{Top-3}   & \textbf{Top-10}   \\ \hline
		SVM-Question                                 & $4.13\pm0.89$    & $9.94\pm1.44$    & $20.91\pm2.08$   &                      & $11.03\pm0.76$    & $21.84\pm0.83$    & $35.89\pm1.31$    &  & $16.38\pm4.38$   & $31.40\pm0.93$   & $50.17\pm2.98$    \\
		SVM-Image                                    & $7.11\pm0.73$    & $19.40\pm2.50$   & $39.75\pm2.23$   &                      & $20.23\pm1.14$    & $33.91\pm1.57$    & $48.15\pm1.18$    &  & $20.95\pm2.17$   & $43.32\pm1.23$   & $57.31\pm2.71$    \\
		SVM-Question+Image                           & $7.35\pm0.73$    & $20.43\pm2.32$   & $40.38\pm2.46$   &                      & $20.76\pm0.95$    & $34.18\pm1.35$    & $48.64\pm1.08$    &  & $21.40\pm2.01$   & $43.33\pm1.45$   & $59.58\pm2.92$    \\
		LSTM-Question                                & $5.08\pm0.54$    & $11.17\pm0.21$    & $23.68\pm2.42$   &                      & $10.96\pm0.60$    & $19.70\pm0.87$    & $32.02\pm0.96$    &  & $16.37\pm2.87$   & $28.31\pm2.11$   & $44.91\pm1.32$    \\
		LSTM-Image                                   & $14.62\pm2.38$   & $29.68\pm4.35$   & $49.74\pm2.56$   &                      & $20.96\pm0.96$    & $36.54\pm1.21$    & $56.32\pm2.31$    &  & $28.82\pm2.79$   & $43.64\pm5.49$   & $62.04\pm3.87$    \\
		{LSTM-Question+Image}                          & {$15.77\pm2.07$}   & {$28.30\pm3.56$}   & {$49.45\pm7.25$}   &                      & {$23.57\pm0.52$}    & {$37.36\pm1.43$}    & {$54.30\pm2.99$}    &  & {$31.74\pm3.69$}   & {$48.18\pm5.80$}   & {$63.59\pm5.60$}    \\
		{LSTM-Question+Image+Pre-VQA}                          & {$15.38\pm1.11$}   & {$32.64\pm2.58$}   & {$51.80\pm3.12$}   &                      & {$25.97\pm0.70$}    & {$41.02\pm1.31$}    & {$57.35\pm1.40$}    &  & {$34.42\pm2.68$}   & {$50.27\pm2.22$}   & {${68.56\pm3.19}$}    \\ 
		{Hie-Question+Image}              & {$27.65\pm1.34$}	& {$45.71\pm1.16$}	& {$62.44\pm1.63$} &      & {$34.00\pm1.33$}	& {$50.01\pm0.53$}	& {$64.05\pm0.46$} &     & {$43.74\pm4.48$}	& {$59.46\pm4.60$}	& {$67.95\pm4.43$}\\
		{Hie-Question+Image+Pre-VQA}		& {$38.39\pm1.33$}	& {$56.07\pm1.05$}	& {$\mathbf{71.60\pm1.82}$} &      & {$43.23\pm0.76$}	& {$59.47\pm0.43$}	& {$\mathbf{72.21\pm0.77}$} &     & {$\mathbf{52.85\pm5.10}$}	& {$\mathbf{66.49\pm3.77}$}	& {$\mathbf{73.40\pm2.35}$}\\
		\hline \hline
		\cellcolor[rgb]{0.8,0.8,0.8}{Ours, gt-QQmaping\textsuperscript{$\ddagger$}}                               & \cellcolor[rgb]{0.8,0.8,0.8}{$65.96\pm2.06$}   & \cellcolor[rgb]{0.8,0.8,0.8}{$78.80\pm1.08$}   & \cellcolor[rgb]{0.8,0.8,0.8}{$79.43\pm1.11$}   & \cellcolor[rgb]{0.8,0.8,0.8}                     & \cellcolor[rgb]{0.8,0.8,0.8}{$64.62\pm0.82$}    & \cellcolor[rgb]{0.8,0.8,0.8}{$71.75\pm0.94$}    & \cellcolor[rgb]{0.8,0.8,0.8}{$73.17\pm0.94$}    &\cellcolor[rgb]{0.8,0.8,0.8}  & \cellcolor[rgb]{0.8,0.8,0.8}{$45.55\pm1.14$}   & \cellcolor[rgb]{0.8,0.8,0.8}{$48.29\pm1.08$}   & \cellcolor[rgb]{0.8,0.8,0.8}{$48.86\pm1.29$}    \\
		Ours, top-1-QQmaping                            & $51.25\pm2.21$   & $63.07\pm1.23$   & $63.13\pm1.21$   &                      & $53.50\pm1.14$    & $60.16\pm0.87$    & $61.20\pm0.96$    &  & $43.54\pm2.14$   & $46.58\pm1.80$   & $47.03\pm2.08$    \\
		Ours, top-3-QQmaping                            & $\mathbf{56.67\pm1.68}$   & $\mathbf{69.31\pm1.10}$   & ${69.36\pm1.03}$   &                      & $\mathbf{57.60\pm1.29}$    & $\mathbf{64.70\pm1.24}$    & ${65.77\pm1.28}$    &  & ${48.74\pm2.47}$    & ${53.45\pm1.99}$   & $53.90\pm2.26$    \\ 
		\hline\hline
		{Ensemble} & {$57.08\pm2.02$} &{-} &{-} & &{$58.98\pm1.04$} &{-} &{-} & & {$59.77\pm5.52$} &{-} &{-}	\\
		\hline\hline
Human & $74.41\pm1.13$
      & -&- & & $78.32\pm0.76$
      &- &- & & $81.95\pm1.83$
      &-&- \\ \hline
	\end{tabular}}
	\caption{Accuracies on the questions that asked based on different Knowledge Base sources. Best single-model results are shown in bold font. $\ddagger$~indicates that ground truth Question-Query mappings are used, which (in \colorbox[rgb]{0.8,0.8,0.8}{gray}) will not participate in rankings.}
	\label{kb-source-table}
\end{table*}

\begin{table*}[t!]
	\centering
	\resizebox{\linewidth}{!}{
	\begin{tabular}{lccclccccccc}
		\hline
		\multicolumn{1}{c}{\multirow{3}{*}{\textbf{Method}}} & \multicolumn{11}{c}{Visual  Concept, Acc. $\pm$ Std (\%)}                                                                                                                                                     \\ \cline{2-12}
		\multicolumn{1}{c}{}                                 & \multicolumn{3}{c}{\textbf{Object}}               & \multicolumn{1}{c}{} & \multicolumn{3}{c}{\textbf{Scene}}                &           & \multicolumn{3}{c}{\textbf{Action}}               \\ \cline{2-4} \cline{6-8} \cline{10-12}
		\multicolumn{1}{c}{}                                 & \textbf{Top-1} & \textbf{Top-3} & \textbf{Top-10} & \textbf{}            & \textbf{Top-1} & \textbf{Top-3} & \textbf{Top-10} & \textbf{} & \textbf{Top-1} & \textbf{Top-3} & \textbf{Top-10} \\ \hline
		SVM-Question                                         & $11.39\pm0.84$          & $22.72\pm0.53$          & $37.89\pm1.06$           &                      & $0.68\pm0.24$           & $1.98\pm0.83$           & $3.97\pm0.74$            &           & $0.00\pm0.00$         & $0.00\pm0.00$           & $0.00\pm0.00$            \\
		SVM-Image                                            & $20.08\pm1.25$          & $35.33\pm1.13$          & $51.60\pm1.11$           &                      & $2.88\pm0.46$           & $5.22\pm0.85$           & $9.60\pm0.36$            &           & $0.00\pm0.00$           & $0.00\pm0.00$           & $0.00\pm0.00$            \\
		SVM-Question+Image                                   & $20.59\pm1.04$          & $35.73\pm1.06$          & $52.24\pm0.79$           &                      & $3.02\pm0.42$           & $5.22\pm0.84$           & $9.83\pm0.19$            &           & $0.00\pm0.00$           & $0.00\pm0.00$           & $0.00\pm0.00$            \\
		LSTM-Question                                        & $11.39\pm0.52$          & $20.70\pm0.63$          & $34.34\pm0.92$           &                      & $1.58\pm1.07$           & $3.05\pm0.67$           & $6.03\pm1.09$           &           & $0.00\pm0.00$           & $3.53\pm4.71$  & $4.71\pm5.76$   \\
		LSTM-Image                                           & $22.34\pm0.98$          & $39.09\pm1.49$          & $60.13\pm2.18$           &                      & $3.80\pm0.92$           & $7.07\pm1.49$          & $14.94\pm1.43$  &           & $0.00\pm0.00$           & $1.05\pm2.11$           & $1.05\pm2.11$            \\
		{LSTM-Question+Image}                                  & {$24.92\pm0.88$}          & {$39.71\pm1.48$}          & {$58.32\pm3.82$}           &                      & {$4.56\pm0.72$}  & {$9.21\pm2.00$}          & {$15.56\pm1.62$}  &           & {$5.99\pm6.64$}           & {$5.99\pm6.64$}  & {$9.52\pm6.15$}   \\
		{LSTM-Question+Image+Pre-VQA}                          & {$26.97\pm0.53$}          & {$43.40\pm0.77$}          & {$61.30\pm0.95$}           &                      & {$6.46\pm1.01$}  & {$12.27\pm1.17$}          & {$19.98\pm2.10$}  &           & {$2.35\pm2.88$}           & {$8.35\pm7.20$}  & {$8.34\pm7.20$}   \\
		{Hie-Question+Image}              & {$36.35\pm1.25$}	& {$53.83\pm1.00$}	& {$68.51\pm0.61$} &      & {$9.09\pm0.46$}	& {$14.36\pm1.30$}	& {$22.54\pm1.27$} &     & {$1.18\pm2.35$}	& {$6.52\pm5.42$}	& {$17.38\pm3.81$}\\
		{Hie-Question+Image+Pre-VQA}		& {$46.38\pm0.81$}	& {$63.65\pm0.55$}	& {$\mathbf{76.81\pm0.48}$} &      & {$12.72\pm0.20$}	& {$\mathbf{19.64\pm1.53}$}	& {$\mathbf{28.92\pm2.74}$} &     & {$8.11\pm8.02$}	& {$19.08\pm5.07$}	& {$21.43\pm3.67$}\\
		\hline \hline
		\cellcolor[rgb]{0.8,0.8,0.8}Ours, gt-QQmaping\textsuperscript{$\ddagger$}                                       & \cellcolor[rgb]{0.8,0.8,0.8}$68.70\pm0.77$          & \cellcolor[rgb]{0.8,0.8,0.8}$76.50\pm0.69$          & \cellcolor[rgb]{0.8,0.8,0.8}$77.11\pm0.71$           &      \cellcolor[rgb]{0.8,0.8,0.8}                & \cellcolor[rgb]{0.8,0.8,0.8}$14.81\pm0.83$          & \cellcolor[rgb]{0.8,0.8,0.8}$20.93\pm1.54$          & \cellcolor[rgb]{0.8,0.8,0.8}$27.75\pm2.03$           &     \cellcolor[rgb]{0.8,0.8,0.8}      & \cellcolor[rgb]{0.8,0.8,0.8}$28.78\pm5.72$           & \cellcolor[rgb]{0.8,0.8,0.8}$39.76\pm6.61$          & \cellcolor[rgb]{0.8,0.8,0.8}$53.32\pm10.47$           \\
		Ours, top-1-QQmaping                                    & $56.75\pm1.48$          & $64.11\pm1.33$          & $64.47\pm1.35$           &                      & $12.81\pm1.29$           & $18.36\pm2.68$          & $24.19\pm3.52$           &           & $16.58\pm10.14$           & $19.57\pm12.38$           & $22.57\pm14.32$            \\
		Ours, top-3-QQmaping                                    & $\mathbf{61.53\pm1.31}$ & $\mathbf{69.51\pm1.35}$ & ${69.90\pm1.20}$  &                      & $\mathbf{12.89\pm1.29}$           & ${18.51\pm2.72}$ & ${24.34\pm3.57}$  &           & $\mathbf{19.75\pm6.16}$           & $\mathbf{22.73\pm8.34}$           & $\mathbf{25.72\pm10.12}$            \\ 
		\hline\hline
		{Ensemble} & {$63.38\pm1.19$} &{-} &{-} & &{$15.13\pm1.24$} &{-} &{-} & & {$14.39\pm3.41$} &{-} &{-}	\\
		\hline\hline
Human & $80.58\pm0.42$
      &- &- & & $52.70\pm2.06$
      &- &- & & $68.23\pm4.90$
      &-&- \\ \hline
\end{tabular}}
	\caption{Accuracies on questions that focus on three different visual concepts.  Best single-model results are shown in bold font. $\ddagger$~indicates that ground truth Question-Query mappings are used, which (in \colorbox[rgb]{0.8,0.8,0.8}{gray}) will not participate in rankings.}
	\label{question_cat}
\end{table*}

\begin{table*}[t!]
	\centering
	\resizebox{0.98\linewidth}{!}
        {
	\begin{tabular}{lccclccc}
		\hline
		\multicolumn{1}{c}{\multirow{3}{*}{\textbf{Method}}} & \multicolumn{7}{c}{Answer-Source, Acc. $\pm$ Std (\%)}                                                                                              \\ \cline{2-8}
		\multicolumn{1}{c}{}                                 & \multicolumn{3}{c}{\textbf{Image}}                & \multicolumn{1}{c}{} & \multicolumn{3}{c}{\textbf{KB}}                   \\ \cline{2-4} \cline{6-8}
		\multicolumn{1}{c}{}                                 & \textbf{Top-1} & \textbf{Top-3} & \textbf{Top-10} & \textbf{}            & \textbf{Top-1} & \textbf{Top-3} & \textbf{Top-10} \\ \hline
		SVM-Question                                         & $12.38\pm0.88$          & $24.68\pm0.47$          & $41.05\pm1.05$           &                      & $0.78\pm0.23$           & $2.00\pm0.47$           & $4.16\pm0.79$            \\
		SVM-Image                                            & $21.81\pm1.30$          & $38.27\pm1.33$          & $55.93\pm1.29$           &                      & $2.30\pm0.25$           & $4.73\pm0.62$           & $7.74\pm0.51$            \\
		SVM-Question+Image                                   & $22.38\pm1.09$          & $38.72\pm1.20$          & $56.63\pm0.97$           &                      & $2.38\pm0.30$           & $4.65\pm0.63$           & $7.86\pm0.44$            \\
		LSTM-Question                                        & $12.35\pm0.57$          & $22.45\pm0.68$          & $37.06\pm0.98$           &                      & $1.45\pm0.68$           & $2.72\pm0.63$           & $5.89\pm0.95$           \\
		LSTM-Image                                           & $24.19\pm0.98$          & $42.40\pm1.73$          & $64.92\pm2.44$           &                      & $3.31\pm0.74$           & $5.69\pm0.87$           & $11.87\pm0.71$           \\
		{LSTM-Question+Image}                                  & {$26.98\pm1.08$}          & {$42.90\pm1.58$}          & {$62.87\pm4.10$}           &                      & {$4.03\pm0.95$}           & {$7.70\pm1.30$}          & {$13.11\pm1.51$}           \\ 
		{LSTM-Question+Image+Pre-VQA}                          & {$28.97\pm0.64$}          & {$46.62\pm0/85$}          & {$65.83\pm0.97$}           &                      & {$6.13\pm1.01$}           & {$10.94\pm1.24$}          & {$16.73\pm1.23$}           \\ 
		{Hie-Question+Image}              & {$39.20\pm1.36$}	& {$57.80\pm1.07$}	& {$73.41\pm0.55$} &      & {$8.16\pm0.39$}	& {$13.81\pm1.46$}	& {$20.78\pm1.11$} \\
		{Hie-Question+Image+Pre-VQA}		& {$49.93\pm1.08$}	& {$68.21\pm0.67$}	& {$\mathbf{82.08\pm0.56}$} &      & {$11.61\pm0.95$}	& {$18.69\pm1.55$}	& {$\mathbf{26.25\pm1.51}$} \\
		\hline \hline
		\cellcolor[rgb]{0.8,0.8,0.8}Ours, gt-QQmaping\textsuperscript{$\ddagger$}                                       & \cellcolor[rgb]{0.8,0.8,0.8}$73.69\pm0.67$          & \cellcolor[rgb]{0.8,0.8,0.8}$81.04\pm0.51$          & \cellcolor[rgb]{0.8,0.8,0.8}$81.04\pm0.51$           & \cellcolor[rgb]{0.8,0.8,0.8}                     & \cellcolor[rgb]{0.8,0.8,0.8}$15.95\pm0.64$           & \cellcolor[rgb]{0.8,0.8,0.8}$25.17\pm0.89$          & \cellcolor[rgb]{0.8,0.8,0.8}$32.39\pm1.16$           \\
		Ours, top-1-QQmaping                                    & $61.11\pm1.48$          & $68.31\pm1.24$          & $68.34\pm1.22$           &                      & $12.12\pm1.01$           & $19.13\pm1.30$          & $23.91\pm1.39$           \\
		Ours, top-3-QQmaping                                    & $\mathbf{66.32\pm1.20}$ & $\mathbf{74.11\pm1.21}$ & ${74.15\pm1.18}$  &                      & $\mathbf{12.39\pm1.09}$  & $\mathbf{19.87\pm1.35}$ & ${24.81\pm1.31}$  \\ 
				\hline\hline
		{Ensemble} & {$68.15\pm0.97$} &{-} &{-} & &{$15.18\pm1.22$} &{-} &{-}	\\
		\hline\hline
Human & $82.97\pm0.38$ &- &- & & $54.47\pm1.77$
      &-&- \\ \hline
	\end{tabular}}
	\caption{Accuracies for different methods according to different answer sources. Best single-model results are shown in bold font. $\ddagger$~indicates that ground truth Question-Query mappings are used, which (in \colorbox[rgb]{0.8,0.8,0.8}{gray}) will not participate in rankings.}
	\label{ans_cource}
\end{table*}

\begin{table*}[t!]
	\resizebox{0.995\linewidth}{!}{
		\begin{tabular}{lllll}
		\multicolumn{2}{c}{\hspace{30pt}
		 \includegraphics[height=3.5cm,width=4.5cm]{1.jpg}}
		&\includegraphics[height=3.5cm,width=4.5cm]{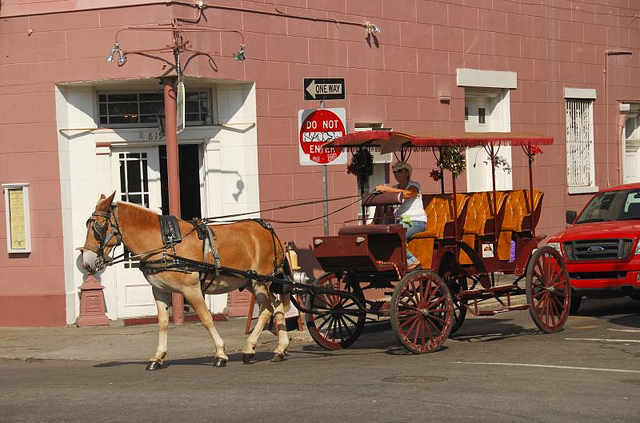}
		&\includegraphics[height=3.5cm,width=4.5cm]{3.jpg}
		&\includegraphics[height=3.5cm,width=4.5cm]{4.jpg}\\
		\multicolumn{2}{l}{\hspace{35pt}
		Which furniture in this image }
		&What animal in this image
		&Which animal in this image
		&Which transportation way in this\\
		\multicolumn{2}{l}{\hspace{35pt}
		can I lie on?}
	    &are pulling carriage?
	    &has stripes?
	    &image is cheaper than taxi?
	    \\ \hline \\ [-2ex] 
		\bluett{\textit{Pred. QT:}} 
		& \bluett{\texttt{(UsedFor,Object,Image)}} 
		& \bluett{\texttt{(CapableOf,Object,Image)} }
		& \bluett{\texttt{(HasA,Object,Image)}}
		& \bluett{\texttt{(Cheaper,Object,Image)}}
		\\ %
		\bluett{\textit{Keywords:}} 
		& \bluett{`lie on'} 
		& \bluett{`pulling carriage'}
		& \bluett{`stripes'}
		& \bluett{`taxi'} \\
		\hline \\ [-2ex] 
		\textit{Pred. SF:} 
		&a sofa is usually to sit or lie on
		&horses sometimes pull carriages 
		&zebras have stripes
		&bus are cheaper than taxi\\
		\textit{Pred. Answer:} &sofa&horse&zebras&bus\\
		\textit{GT Answer:} &sofa&horse&zebras&bus\\ \hline \\ [-2ex] 
		\multicolumn{2}{c}{\hspace{30pt}
		 \includegraphics[height=3.5cm,width=4.5cm]{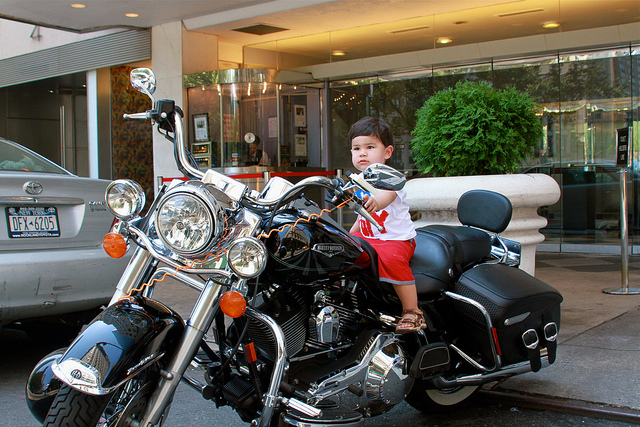}}
		&\includegraphics[height=3.5cm,width=4.5cm]{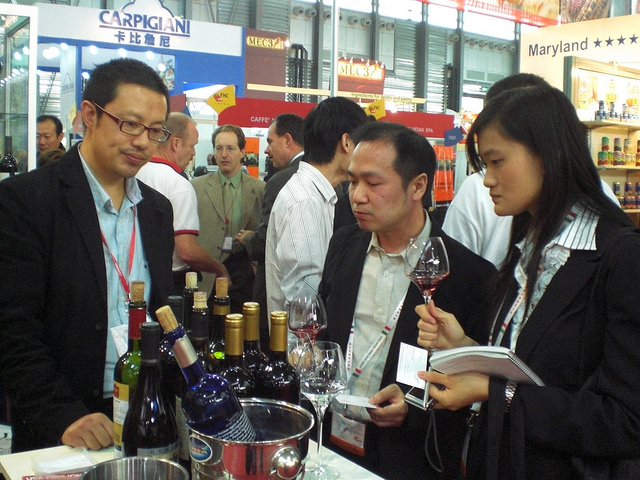}
		&\includegraphics[height=3.5cm,width=4.5cm]{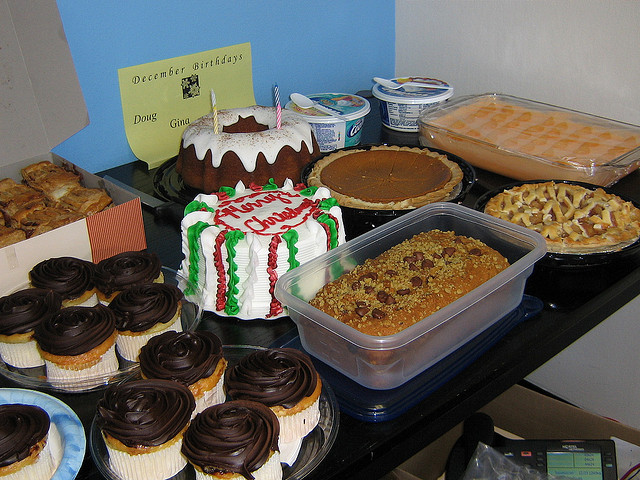}
		&\includegraphics[height=3.5cm,width=4.5cm]{8.jpg}\\
		\multicolumn{2}{l}{\hspace{35pt}
		 Which object in this image}
		&What in this image is 
		&Which food in this image can 
		&What animal can be found in \\
		\multicolumn{2}{l}{\hspace{35pt}
		 can I ride?}
	    &helpful for a romantic dinner?
	    &be seen on a birthday party?
	    &this place?\\ \hline \\ [-2ex] 
		\bluett{\textit{Pred. QT:}} & \bluett{\texttt{(UsedFor,Object,Image)} }
		& \bluett{\texttt{(HasProperty,Object,Image)} }
		& \bluett{\texttt{(AtLocation,Object,Image)}}
		& \bluett{\texttt{(AtLocation,Scene,KB)}}
		\\ %
		\bluett{\textit{Keywords:}} & \bluett{`ride'} 
		& \bluett{`romantic dinner'}
		& \bluett{`birthday party'}
		& \bluett{-} \\
		\hline	\\ [-2ex] 			
		\textit{Pred. SF:} 
		&motorcycle is used for riding
		&wine is good for a romantic dinner 
		&cake is related to birthday party
		&You are likely to find a cow \\
		&
		&
		&
		&in a pasture\\
		\textit{Pred. Answer:} &motorcycle&wine&cake&cow\\
		\textit{GT Answer:} &motorcycle&wine&cake&cow\\ \hline \\ [-2ex] 
		\multicolumn{2}{c}{\hspace{30pt}
		 \includegraphics[height=3.5cm,width=4.5cm]{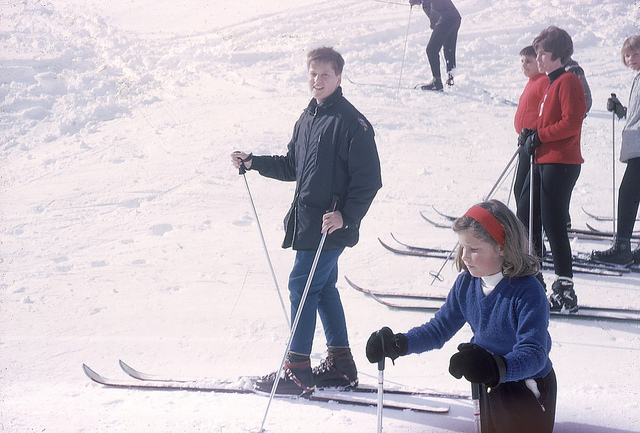}}
		&\includegraphics[height=3.5cm,width=4.5cm]{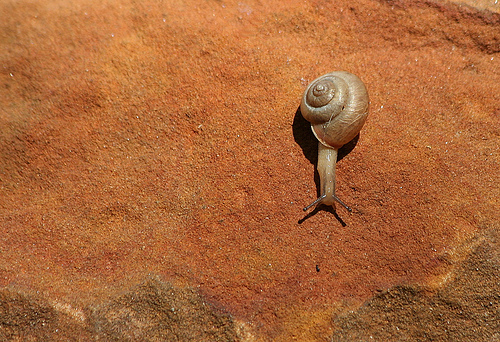}
		&\includegraphics[height=3.5cm,width=4.5cm]{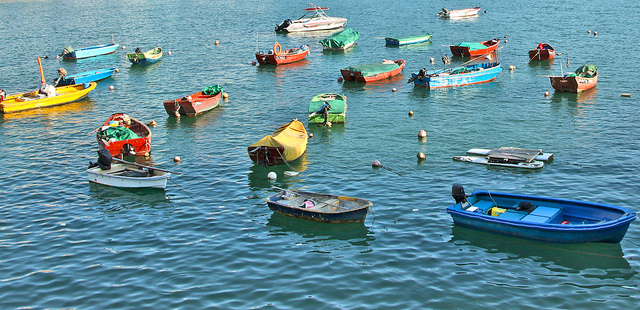}
		&\includegraphics[height=3.5cm,width=4.5cm]{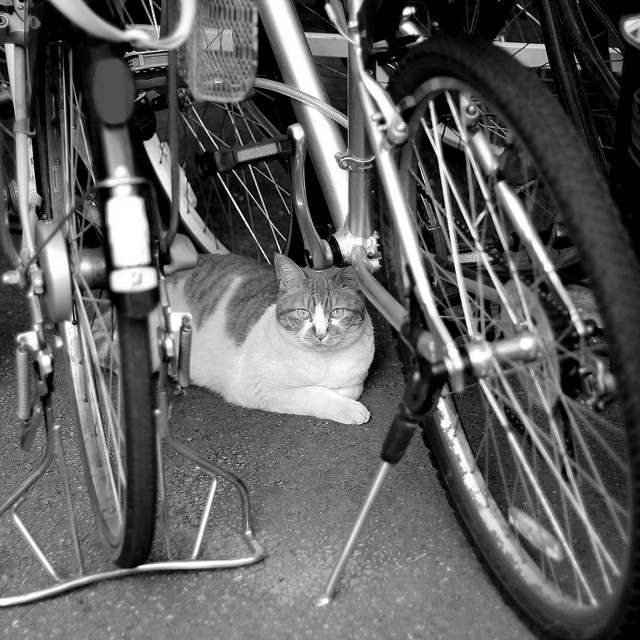}\\
		\multicolumn{2}{l}{\hspace{35pt}
		 What kind of people can we}
		&What does the animal in the right 
		&Which object in this image is 
		&What in this image is  \\
		\multicolumn{2}{l}{\hspace{35pt}
		usually find in this place?}
	    &of this image have as a part?
	    &related to sail? 
	    &capable of hunting a mouse?\\ \hline \\ [-2ex] 
		\bluett{\textit{Pred. QT:}} & \bluett{\texttt{(AtLocation,Scene,KB)}} 
		& \bluett{\texttt{(PartOf,Object,KB)} }
		& \bluett{\texttt{(RelatedTo,Object,Image)}}
		& \bluett{\texttt{(CapableOf,Object,Image)}}
		\\ %
		\bluett{\textit{Keywords:}} & \bluett{-}
		& \bluett{-}
		& \bluett{`sail'}
		& \bluett{`hunting mouse'} \\
		\hline \\ [-2ex] 
		\textit{Pred. SF:} 
		&skiiers can be on a ski slope
		&snails have shells
		&boat is related to sailing
		&a cat can hunt mice \\
		\textit{Pred. Answer:} &skiiers&shells&boat&cat\\
		\textit{GT Answer:} &skiiers&shells&boat&cat\\ \hline \\ [-2ex] 
		\multicolumn{2}{c}{\hspace{30pt}
   		 \includegraphics[height=5cm,width=4.5cm]{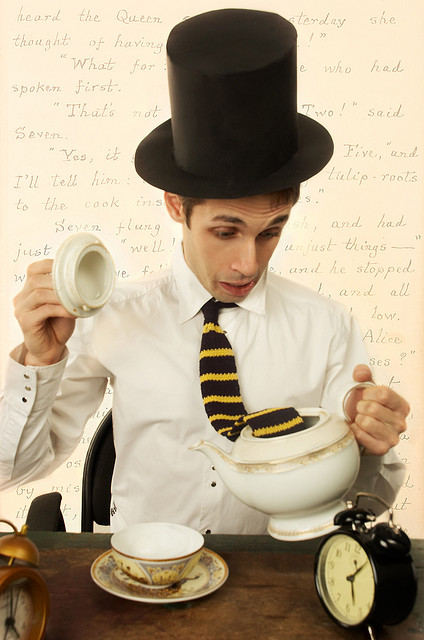}}
		&\includegraphics[height=5cm,width=4.5cm]{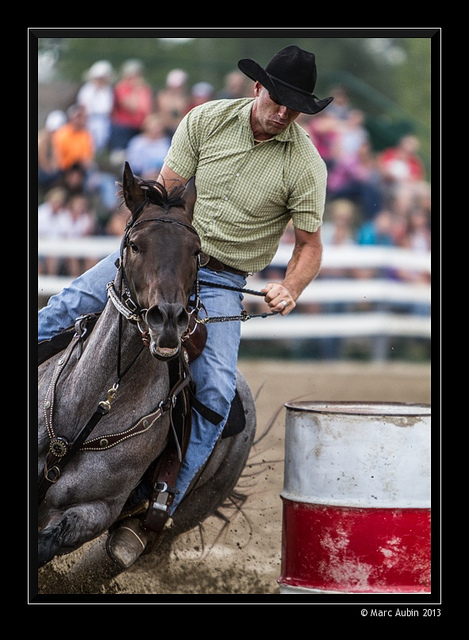}
		&\includegraphics[height=5cm,width=4.5cm]{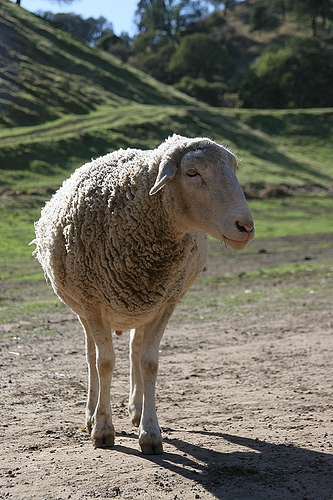}
		&\includegraphics[height=5cm,width=4.5cm]{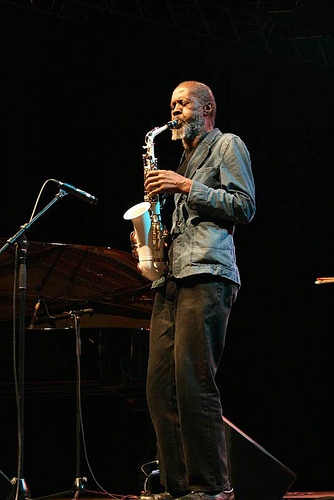}\\
		\multicolumn{2}{l}{\hspace{35pt}
			Which object in this image is used}
		&Which object in this image 
		&Which object in this image 
		&Which instrument in this   \\
		\multicolumn{2}{l}{\hspace{35pt} 
			to measure the passage of time?}
		&is a very trainable animal?
		&is related to wool?
		&image is common in jazz?\\ \hline \\ [-2ex] 
		\bluett{\textit{Pred. QT:}} & \bluett{\texttt{(UsedFor,Object,Image)} }
		& \bluett{\texttt{(IsA,Object,Image)} }
		& \bluett{\texttt{(RelatedTo,Object,Image)}}
		& \bluett{\texttt{(IsA,Object,Image)}}
		\\ %
		\bluett{\textit{Keywords:}} & \bluett{`measure passage time'}
		& \bluett{`trainable animal'}
		& \bluett{`wool'}
		& \bluett{`jazz'} \\
		\hline \\ [-2ex] 
		\textit{Pred. SF:} 
		&a clock is for measuring the 
		&horses are very trainable animals
		&sheep is related to wool
		&a saxophone is a common instrument\\
		&passage  of time
		&
		&
		&in jazz\\
		\textit{Pred. Answer:} &clock&horse&sheep&saxophone\\
		\textit{GT Answer:} &clock&horse&sheep&saxophone\\ \hline
   \end{tabular}}
	\caption{Some example results generated by our methods (Pred.: predicted, 
		QT: Question Type, SF: supporting-fact, GT: ground truth). 
		The question type is represented by a $3$-turple (\texttt{$T_{\mbox{REL}}$}, \texttt{$T_{\mbox{KVC}}$}, \texttt{$T_{\mbox{AS}}$}).
		The supporting-facts triplet have been translated to textual sentence for easy understanding.
	    Note that no keywords are mined by our approach, if the predicted answer-source (\texttt{AS}) is \texttt{KB}.}
	\label{results_examples}
	\vspace{1cm}
\end{table*}

\begin{table*}[t!]
	\resizebox{\linewidth}{!}{
		\begin{tabular}{lllll}
			\multicolumn{2}{c}{\hspace{40pt}
				         \includegraphics[height=4cm,width=4.5cm]{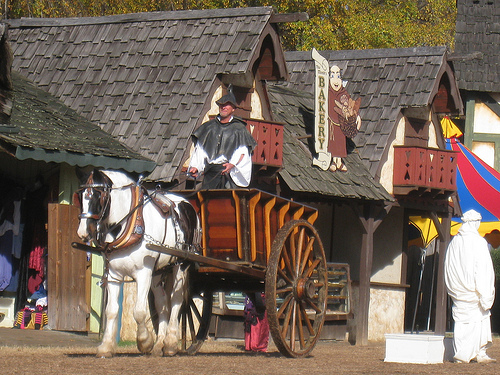}}
                        &\includegraphics[height=4cm,width=4.5cm]{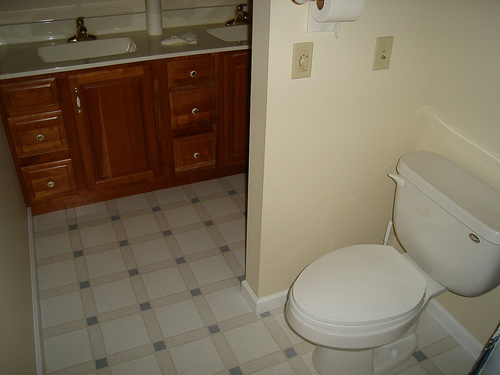}
                        &\includegraphics[height=4cm,width=4.5cm]{COCO_val2014_000000000802.jpg}
                        &\includegraphics[height=4cm,width=4.5cm]{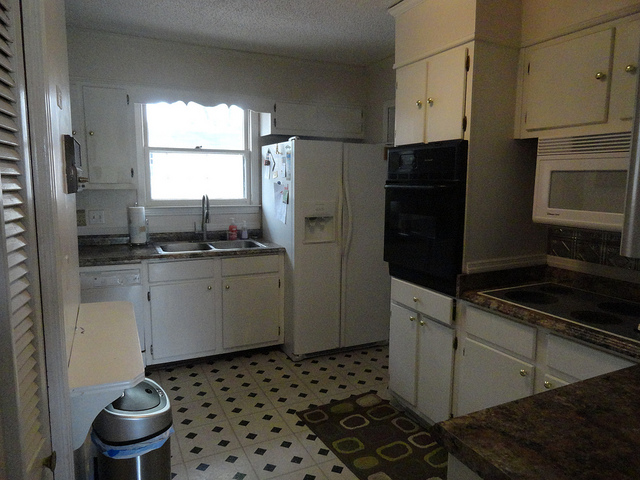}\\
			\multicolumn{2}{l}{\hspace{44pt}What animal in this image can }
                        & What can the place in the image
                        & Which object in this image
                        & What can I do using this place?\\
			\multicolumn{2}{l}{\hspace{44pt}rest standing up?}
                        & be used for?
                        & is utilized to chill food?
                        &\\ \hline \\ [-2ex] 
			\textit{Pred. VC:}
                        & Person, Cart, ...
                        & Kitchen, ...
                        & Refrigerator, Over, Stove, ...
                        & Kitchen, Refrigerator, ...\\
			\textit{GT VC:}
                        & Horse
                        & Bathroom
                        & Refrigerator
                        & Kitchen\\ \hline \\ [-2ex] 
			\textit{Pred. QT:}
                        & \texttt{(CapableOf,Object,Image,)}
                        & \texttt{(UsedFor,Scene,KB)}
                        & \texttt{(IsA,Object,Image)}
                        & \texttt{(UsedFor,Scene,KB)}\\
			\textit{GT QT:}
                        & \texttt{(CapableOf,Object,Image)}
                        & \texttt{(UsedFor,Scene,KB)}
                        & \texttt{(UsedFor,Object,Image)}
                        & \texttt{(UsedFor,Scene,KB)}\\ \hline \\ [-2ex] 
			\textit{Pred. SF:}
                        & People can stand up for themselves
                        & A bathroom is for washing your hands
                        & An oven is a device to heat food
                        & A kitchenette is for cooking\\
			\textit{GT SF:}
                        & Horses can rest standing up
                        & A kitchen is for cooking
                        & A refrigerator is used for chilling food
                        & A kitchenette is for preparing food\\ \hline \\ [-2ex] 
			\textit{Pred. Answer:}
                        & People
                        & Cooking
                        & Oven
                        & Cooking\\
			\textit{GT Answer:}
                        & Horse
                        & Washing
                        & Refrigerator
                        & Preparing food\\ \hline
		\end{tabular}}
	        \caption{Failure cases of our approach (Pred.: predicted, GT: ground truth, QT: query type, VC: visual concept, SF: supporting-fact).
	        	The question type is represented by a $3$-turple (\texttt{$T_{\mbox{REL}}$}, \texttt{$T_{\mbox{KVC}}$}, \texttt{$T_{\mbox{AS}}$}).
                          The false reason for the first two examples is that the visual concepts are not extracted correctly.
                          Our method makes a mistake on the third example due to the false question-to-query mapping.
                          The reason for the fourth example is that the question has multiple answers (our method orders these answers according to the frequency in the training data, see Section~\ref{sec:answering} for details).
                          }
		\label{neg_results_examples}
\end{table*}

Table \ref{over-acc} shows the overall accuracy of all the baseline methods and our proposed models. In the case of Top-$1$ accuracy, our `{top-3-QQmaping}' is the best-performing single model, which doubles the accuracy of the baseline `LSTM-Question+Image'. {Our proposed model also outperforms `LSTM-Question+Image+Pre-VQA' and `Hie-Question+Image+Pre-VQA', although the latter use additional training data. The `Ensemble' model achieves the best accuracy of $58.76\%$, showing that the conventional LSTM-based model and our knowledge based model are complementary.} The `top-3-QQmaping model' outperforms `top-1-QQmaping' because it produces better Question-Query mapping results, as shown in the Table \ref{question_query_map}. However, it is still not as good as `gt-QQmaping' due to Question-Query mapping errors. There is still a significant gap between our models and the human performance. Among the baseline models, LSTM methods perform slightly better than SVM. `Question+Image' models always predict more accurate answers than `Question' or `Image' alone, no matter in SVM or LSTM. Interestingly, contradictory with previous works \cite{ren2015image,antol2015vqa} which found that `question' plays a more important role than `image' in the VQA \cite{antol2015vqa} or COCO-QA datasets~\cite{ren2015image}, our `\{SVM,LSTM\}-Q' performs worse than `\{SVM,LSTM\}-I', meaning that FVQA questions rely more heavily on the image content than the existing VQA datasets. Actually, if `\{SVM,LSTM\}-Q' achieves too high performance, the corresponding questions may be not \textit{Visual Questions} and they may be actually \textit{Textual Questions}. { The LSTM models that are pretrained on the VQA dataset performs slightly better than training from scratch, but not as well as our models. Using the Top-$3$ measure, our `{top-3-QQmaping}' model also performs best. We produce slightly lower Top-$10$ performance, because our proposed methods may produce less than $10$ answers in some cases.}

Table~\ref{results_examples} shows some example results generated by our final model. \bluetext{Tables~\ref{over-WUPS9} and \ref{over-WUPS0} report the WUPS@0.9 and WUPS@0.0 accuracy for different methods.}

Table~\ref{kb-source-table} reports the accuracy for different Knowledge Base sources.
Our `top-3-QQmapping' model performs better than other baselines for DBpedia and ConceptNet KBs. 
For Webchild, `top-3-QQmapping' is only worse than the state-of-the-art `Hie-Question+Image+Pre-VQA' model, which, however, uses extra data.

\begin{table*}[h!]
\centering
\resizebox{.6\linewidth}{!}{
\begin{tabular}{lccc}
\hline
\multicolumn{1}{c}{\multirow{2}{*}{\textbf{Method}}} & \multicolumn{3}{c}{Supporting-Fact Prediction Acc. $\pm$ Std (\%)}     \\ \cline{2-4}
\multicolumn{1}{c}{}                                 & \textbf{Top-1} & \textbf{Top-3} & \textbf{Top-10} \\ \hline
\cellcolor[rgb]{0.8,0.8,0.8}Ours, gt-QQmaping\textsuperscript{$\ddagger$}                                       & \cellcolor[rgb]{0.8,0.8,0.8}$56.31\pm0.89$          & \cellcolor[rgb]{0.8,0.8,0.8}$62.55\pm0.96$          & \cellcolor[rgb]{0.8,0.8,0.8}$63.55\pm0.96$           \\
Ours, top-1-QQmaping                                    & $38.76\pm0.88$          & $42.96\pm0.78$          & $43.60\pm0.77$           \\
Ours, top-3-QQmaping                                    & $\mathbf{41.12\pm0.74}$ & $\mathbf{45.49\pm0.89}$ & $\mathbf{46.13\pm0.87}$  \\ \hline
\end{tabular}}
\caption{Facts prediction accuracy for our proposed methods. Best results are shown in bold font. $\ddagger$~indicates that ground truth Question-Query mappings are used, which (in \colorbox[rgb]{0.8,0.8,0.8}{gray}) will not participate in rankings.}
\label{facts}
\end{table*}

Table~\ref{question_cat} illustrates the performance on questions that focus on different types of visual concepts, which are object, scene and action.
The performance on object-related questions is significantly higher than the other two types, especially when image features are given.
This is not surprising since image features are extracted from the VggNet which has been pre-trained on the object classification task.
{
The accuracy of action or scene related questions is poorer than object-related questions (even for human subjects),
which is partially because the answers of many scene or action related questions can be expressed in different ways.
For example, the answer to `What can I do in this place' (the image scene is kitchen) can be `preparing food' or `cooking'.
On the other hand, the performance of action classification is also worse than objects, which also leads to poor VQA performance.
}

Table~\ref{ans_cource} presents the accuracy for different methods according to different Answer Sources. If the answer is a visual concept in the image, we categorize the answer source into `Image'. Otherwise, it is categorized into `KB'. From the table, we can see that the accuracy is much higher when the answer is from the `Image' side, nearly $5$ times as much as `KB'. This suggests that generating answers from a nearly unlimited answer space (and the answer is not directly appeared in the image) is a very challenging task. Our proposed models performs better than other baseline models.

{
Table \ref{results_examples} shows some examples in which our method (`top-3-QQmapping') achieves the correct answer and Table~\ref{neg_results_examples}
shows some failure cases of our approach.
From Table~\ref{neg_results_examples}, we can see that the failure reasons are categorized into three aspects:
1. The visual concepts of the input image are not extracted correctly. In particular, the errors usually occur when the questioned visual concepts are missing.
2. The question-to-query mapping (via LSTM) is not correct, which means that the question text is wrongly understood.
3. Some errors occur during the stage of post-processing that generates the final answer from queried KB facts.
The approach should select the most relevant fact from multiple facts that matches query conditions.
In particular for questions whose answers are from KBs (in order words, open-ended questions), our method may generate multiple answers.
Sometimes, the ground truth is not the first in the ordered answers. In these cases, the top$1$ answer is wrong, but the top$N$ answers may be correct.
}

Different from all the other state-of-art VQA methods, our proposed models are capable of explicit reasoning, \ie, providing the supporting-facts of predicted answers. Table~\ref{facts} reports the accuracy of supporting-fact prediction. We have $41\%$ chance to predict the correct one from millions of facts in the incorporated Knowledge Bases.

\section{Conclusion}
\label{conclusion}

In this work, we have built a new dataset and an approach for the task of visual question answering with external commonsense knowledge.
The proposed \KBName dataset differs from existing VQA datasets in that it provides a supporting-fact which is critical for answering each visual question.
We have also developed a novel VQA approach, which is able to automatically find the supporting-fact
for a visual question from large-scale structured knowledge bases.
Instead of directly learning the mapping from questions to answers, our approach learns the mapping from questions to KB-queries,
so it is much more scalable to the diversity of answers.
Not only give the answer to a visual question, the proposed method also provides the supporting-fact based on which it arrives at the answer,
which uncovers the reasoning process.

{
\bibliographystyle{IEEEtran}
\bibliography{CSRef}
}

\begin{IEEEbiographynophoto}{Peng Wang}
	is a Professor at Northwestern Polytechnical University, China. He was with 
	the Australian Centre for Visual Technologies (ACVT) of the University of Adelaide from 2012 to 2017. He received a Bachelor in electrical engineering and automation, and a PhD in control science and engineering from Beihang University (China) in 2004 and 2011, respectively.
\end{IEEEbiographynophoto}

\begin{IEEEbiographynophoto}{Qi Wu}
	is a postdoctoral researcher at the University of Adelaide. His research interests include cross-depiction object detection and classification, attributes learning, neural networks, and image captioning. He received a Bachelor in mathematical sciences from China Jiliang University, a Masters in Computer Science, and a PhD in computer vision from the University of Bath (UK) in 2012 and 2015, respectively.
\end{IEEEbiographynophoto}

\begin{IEEEbiographynophoto}{Chunhua Shen}
	is a Professor of computer science at the University of Adelaide. He was with the computer vision program at NICTA (National ICT Australia) in Canberra for six years before moving back to Adelaide. He studied at Nanjing University (China), at the Australian National University, and received his PhD degree from the University of Adelaide. In 2012, he was awarded the Australian Research Council Future Fellowship.
\end{IEEEbiographynophoto}

\begin{IEEEbiographynophoto}{Anthony Dick}
	is an Associate Professor at the University of Adelaide. He
	received a PhD degree from the University of Cambridge in 2002, where he worked on 3D reconstruction of architecture from images. His research interests include image-based modeling, automated video surveillance, and image search.
\end{IEEEbiographynophoto}

\begin{IEEEbiographynophoto}{Anton van den Hengel}
	is a Professor at the University of Adelaide and the founding Director of The Australian Centre for Visual Technologies (ACVT). He received a PhD in Computer Vision in 2000, a Master Degree in Computer Science in 1994, a Bachelor of Laws in 1993, and a Bachelor of Mathematical Science in 1991, all from The University of Adelaide.
\end{IEEEbiographynophoto}

\end{document}